\documentclass[a4paper,twoside,pdftex]{article}

\usepackage[pdftex]{graphicx}  % Load graphicx with the correct options first
\usepackage{epsfig}
\usepackage{subcaption}
\usepackage{calc}
\usepackage{amssymb}
\usepackage{amstext}
\usepackage{amsmath}
\usepackage{amsthm}
\usepackage{multicol}
\usepackage{pslatex}
\usepackage{apalike}
\usepackage{xcolor}
\definecolor{newcolor}{rgb}{.8,.349,.1}

\usepackage{url}
\usepackage{ulem}
\usepackage[utf8]{inputenc}  % Add this line to handle Unicode characters
\usepackage{hhline}
\usepackage{array}
\usepackage[super]{nth} 
\usepackage{multirow}
\usepackage{framed}
\usepackage{SCITEPRESS}  % Please add other packages that you may need BEFORE the SCITEPRESS.sty package.

\begin{document}

% \title{AttentNet for Pulmonary Lung Nodule Detection Using 3D Attention}
% \title{AttentNet: 3D Convolutional Attention for Pulmonary Nodule Detection}
\title{AttentNet: Fully Convolutional 3D Attention for Lung Nodule Detection}

% A.P. ORCID :    0000-0001-5114-1514
% X.X. ORCID :    0000-0002-2701-8660
% M.M. ORCID :    0000-0002-5748-1760
% J.A. ORCID :    0000-0002-0156-8162

% \thanks{Corresponding author}
\author{\authorname{Majedaldein Almahasneh\textsuperscript{*}
% \thanks{Corresponding author}
\sup{1}\orcidAuthor{0000-0002-5748-1760},
Xianghua Xie\sup{2}\orcidAuthor{0000-0002-2701-8660},
and Adeline Paiement\sup{3}\orcidAuthor{0000-0001-5114-1514}}
\affiliation{\sup{1}Department of Computer Science, Loughborough University, Loughborough, UK}
\affiliation{\sup{2}Department of Computer Science, Swansea University, Swansea, UK}
\affiliation{\sup{3}Université de Toulon, Aix Marseille Univ, CNRS, LIS, Marseille, France}}


\abstract{
Motivated by the increasing popularity of attention mechanisms, we observe that popular convolutional (conv.) attention models like Squeeze-and-Excite (SE) and Convolutional Block Attention Module (CBAM) rely on expensive multi-layer perception (MLP) layers. These MLP layers significantly increase computational complexity, making such models less applicable to 3D image contexts, where data dimensionality and computational costs are higher. In 3D medical imaging, such as 3D pulmonary CT scans, efficient processing is crucial due to the large data volume. Traditional 2D attention generalized to 3D increases the computational load, creating demand for more efficient attention mechanisms for 3D tasks. We investigate the possibility of incorporating fully convolutional (conv.) attention in 3D context. We present two 3D fully conv. attention blocks, demonstrating their effectiveness in 3D context. Using pulmonary CT scans for 3D lung nodule detection, we present AttentNet, an automated lung nodule detection framework from CT images, performing detection as an ensemble of two stages, candidate proposal and false positive (FP) reduction. We compare the proposed 3D attention blocks to popular 2D conv. attention methods generalized to 3D modules and to self-attention units. For the FP reduction stage, we also use a joint analysis approach to aggregate spatial information from different contextual levels. We use LUNA-16 lung nodule detection dataset to demonstrate the benefits of the proposed fully conv. attention blocks compared to baseline popular lung nodule detection methods when no attention is used. Our work does not aim at achieving state-of-the-art results in the lung nodule detection task, rather to demonstrate the benefits of incorporating fully conv. attention within a 3D context.
}

\onecolumn \maketitle \normalsize \setcounter{footnote}{0} \vfill

\section{Introduction}
\label{section:introduction}

\renewcommand{\thefootnote}{\fnsymbol{footnote}} % Change footnote marker to symbols
\setcounter{footnote}{1} % Set footnote counter to asterisk
\footnotetext{corresponding author.}

The strength of CNNs manifests in their rich representational power and ability in embedding spatial and cross-channel information. However, due to their inherent locality, CNNs suffer in modeling long-range relations. 
Moreover, performance of CNNs may be prone to network factors, such as depth and width 
% \cite{kopuklu2019resource,method:resnext,method:resnet,szegedy2015going,zagoruyko2016wide}
\cite{kopuklu2019resource,method:resnext,method:resnet}, and data characteristics such as target structures with high morphological variance (e.g. texture, shape, and size). 
To combat these limitations and enhance the performance of CNNs, research efforts has been directed towards attention mechanisms \cite{woo2018cbam,hu2018squeeze,dosovitskiy2020image}.
% \cite{woo2018cbam,hu2018squeeze,schlemper2019attention,dosovitskiy2020image}.
Attention simulates the cognitive process of selective focus on features with high relevance to a task while excluding others.
Incorporating such methods can elevate CNNs performance by automatically learning to focus on important structures associated with a given objective.

The introduction of data collections (e.g. LUNA16 dataset \cite{dataset:LUNA16}) attracted more interest to the pulmonary nodule detection problem. It also made the integration of DL tools possible, however, training robust DNNs requires huge amounts of labeled samples, which remains a challenge when dealing with pulmonary images. Indeed, preparing such datasets is very resource intensive due to the 3D nature of the CT images and the high complexity of the task.
Moreover, available DL methods are designed for generic object detection tasks and typically aim at analysing 2D and RGB images. They are not suited to be directly applied to the medical imaging domain (e.g. 3D CT images) due to the different nature of the data and the complexity of the task.
Additionally, unlike common computer vision problems, the sparse nature of lung nodules within the lung region, along with the prevailing class imbalance in the pulmonary data, make the detection task more challenging.
Therefore, this scenario requires designing a specialised DL approach.

Computer aided detection (CAD) are systems designed to assist radiologists in detecting lung nodules. Incorporating such systems has been clinically proven to decrease observational oversights (i.e. false negative rate) while significantly reducing the reading time required per scan and retaining a consistent quality
% \cite{castellino2005computer,matsumoto2013computer,al2017review}
\cite{matsumoto2013computer,al2017review}. Indeed, a number of studies have demonstrated that CAD systems were able to detect nodules that were originally missed by experienced radiologists \cite{jacobs2016computer,al2017review}. 
% \cite{jacobs2016computer,armato2002lung,yuan2006computer,lee2005lung,al2017review}. 
% 

Generally, CAD systems consists of two consecutive stages, a candidate proposal stage, in which candidate locations are proposed 
at a high
sensitivity and typically on the account of high 
false positive rates, and a false positive reduction stage
to minimise the number of the false alarms and produce the final set of predictions \cite{dataset:LUNA16}.

% \sout{Thanks to Deep Learning (DL), object detection has evolved drastically in the past two decades.
% The emergence of convolutional neural networks (CNN) has transformed object detection from handcrafted low-level feature (e.g. Haar
% % \cite{paper:violaandjones}
% , and HOG
% % \cite{paper:hogfeatures}
% ) based learning to a high-level and semantic feature learning task.}

Several deep neural networks (DNN) has been proposed for generic object detection (e.g. YOLO
% \cite{method:yolo}
, SSD
% \cite{method:SSD}
, R-FCN 
% \cite{paper:RFCN}
, Cornernet 
% \cite{paper:cornernet}
, and Faster RCNN 
% \cite{ren2015faster}
).
These may be divided into two groups:
two stage detectors, in which objects are detected in two steps, region proposal and classification, and single stage methods, where a DNN is trained to regress object locations and classes simultaneously. In general, two stage detectors (e.g. Faster RCNN) can achieve higher accuracy in contrast to single stage detectors \cite{paper:opttradeoff,Huang_2017}.

In this work, we investigate the possibilities of incorporating DL methods to solve the pulmonary detection problem. Specifically, we 
% \sout{present }
utilise a two stage (candidate proposal and false positive reduction) detector based on Faster RCNN \cite{ren2015faster}. We adopt a 3D encoder-decoder structure for the backbone network of our candidate proposal stage. 
Moreover, motivated by the high relevance of nodule morphology and the sparse nature of nodule locations, as well as the success of attention mechanisms in medical imaging related tasks (e.g. detection
% \cite{xiao2021pam,lu2021multi,nawshad2021attention,sangeroki2021fast,sun2021attention}
\cite{lu2021multi,nawshad2021attention,sangeroki2021fast,sun2021attention}
and segmentation 
% \cite{fu2021multimodal,chen2021transunet,wang2021transbts,zheng2021rethinking})
\cite{chen2021transunet,wang2021transbts,zheng2021rethinking}), %
we evaluate different state-of-the-art attention mechanisms and propose two fully convolutional attention blocks that demonstrate potential within the pulmonary nodule detection task.
We also utilise a cross-sectional augmentation approach to battle the limited availability of annotated samples.
Additionally, in line with the objective of the candidate proposal stage, we exploit a testing time augmentation strategy that further enhances the sensitivity of the trained model.
To tackle the class imbalance problem present in the lung data, we adopt focal loss \cite{lin2017focal} and exploit an online hard example mining technique \cite{method:OHEM}.

\begin{figure}
	\centering
	\includegraphics[width=\linewidth]{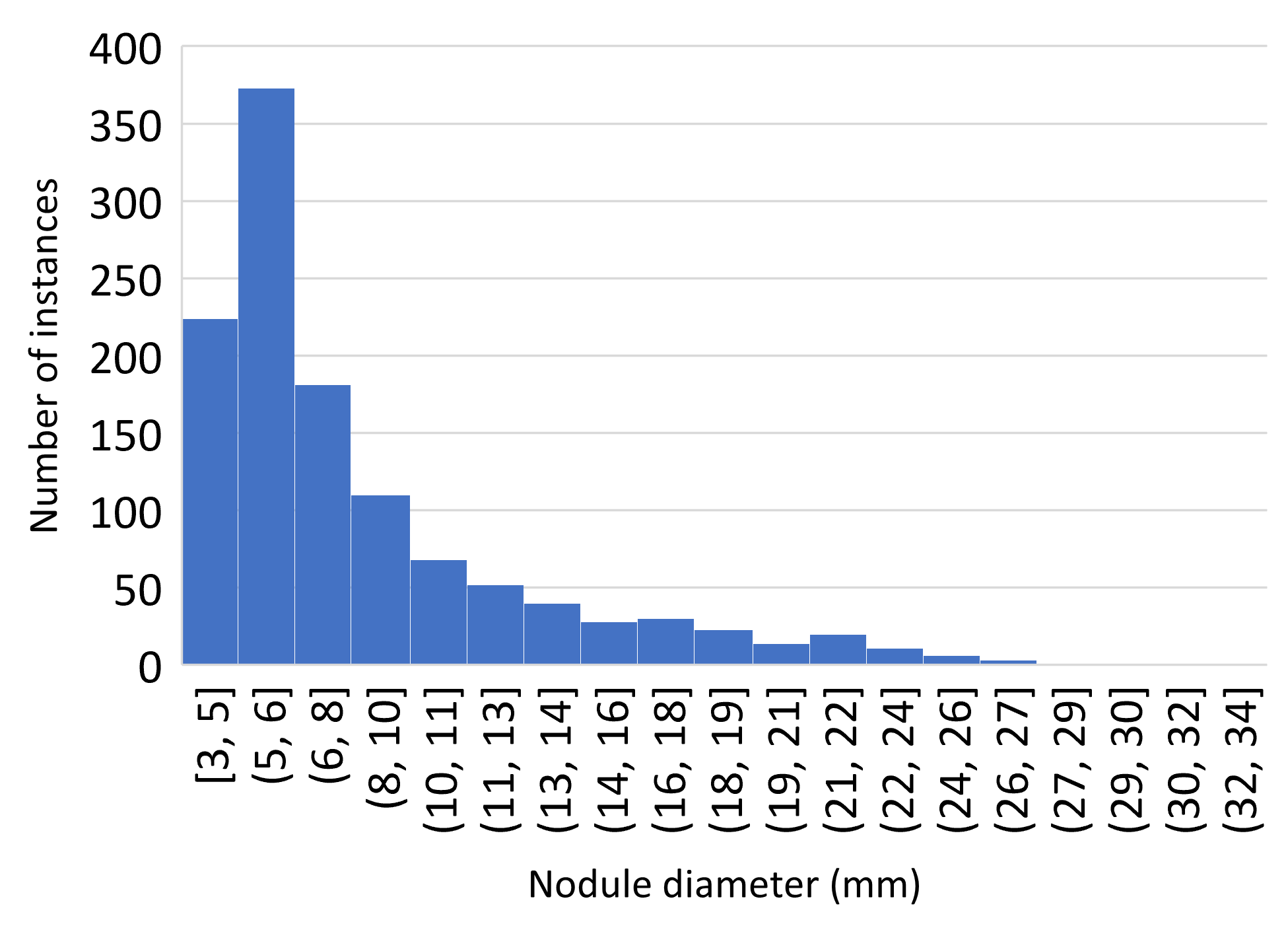}
	\caption{Distribution of nodule diameters in LUNA16 \cite{dataset:LUNA16} dataset. The average nodule diameter is 8.32 mm.}
	\label{fig:nodule_size_distribution_luna16}
\end{figure}

For our False positive reduction stage, in addition to exploiting attention techniques, and inspired by the variant nodule sizes (See Fig. \ref{fig:nodule_size_distribution_luna16}), we adopt a joint analysis based approach in which information from different contextual levels is aggregated and analysed simultaneously. We also propose a zoom-in convolutional path to assist the network in capturing multi-scale spatial embeddings.

% \sout{
% Lastly, we propose a modified version of ReLU \cite{glorot2011deep} activation in an attempt to refine it against dying neurons by
% allowing the generation of small negative outputs for inputs that lie in the flat segment of ReLU.
% We empirically demonstrate the benefits of our proposed activation on within the lung nodule detection task.}

More formally, our contributions may be summarised as follows:
\begin{enumerate}
    % \item We present a framework to handle 3D pulmonary nodule detection from CT images. Our framework detects nodules in two stages, candidate proposal and false positive reduction. We evaluate our framework on the publicly available dataset, LUNA16, and demonstrate an outstanding performance against 
    % % state-of-the-art
    % competing lung nodule detection methods. 

    % \item We propose two fully convolutional attention blocks and demonstrate their potential within the lung nodule detection task. We also carry out extensive experiments using different state-of-the-art attention mechanisms and compare their performance against our results. 
    
    % You could detail your contributions more, by highlighting the idea of using different orientations in the spatial attention module.
    \item We propose two fully convolutional attention blocks 
    in which we incorporate 3D features to infer cross-channel and cross-sectional spatial
    correlations.
    We demonstrate the potential of our attention approach within the lung nodule detection task.
    We also carry out extensive experiments using different state-of-the-art attention mechanisms and compare their performance against our proposed methods. 
    % We present a framework to handle 3D pulmonary nodule detection from CT images. Our
    We evaluate our attention blocks within a 3D lung nodule detection framework that detects nodules in two stages, candidate proposal and false positive reduction. 
    % We evaluate our framework on the publicly available dataset, LUNA16, to demonstrate the benefits of our attention blocks
    % demonstrate an 
    % outstanding 
    We use on the publicly available dataset, LUNA16, and compare the performance against existing lung nodule detection baseline methods where no attention is used.
    % to demonstrate the benefits of our attention blocks
    % performance against 
    % state-of-the-art
    % competing lung nodule detection methods. 

    % \item We present a testing time cross-sectional augmentation strategy that further enhances the performance of the candidate proposal stage.

    \item We employ a joint analysis based approach that elevates the performance of the false positive reduction stage by exploiting different levels of spatial contextual information simultaneously. We present a zoom-in convolutional block that allows the network learning information from different scales and therefore enhance the final prediction.
    
    % \item We propose a modification of the ReLU activation function to reduce the risk of the dying ReLU problem and empirically evaluate its influence within the lung nodule detection task.
\end{enumerate}

In the remainder of this 
% chapter
article,
Section \ref{section:relatedwork} discusses related literature.
Section \ref{section:methodology} presents the details of our proposed framework, 
Section \ref{section:experiments} discusses datasets, pre-processing, experiments and results.
Last, Section \ref{section:conclusion} concludes our work.

% Distribution of nodule diameters in LUNA16

\section{Related work}
\label{section:relatedwork}

\subsection{Attention-mechanisms}

Attention in DL may be split into three main categories, channel-wise, spatial-wise, and global attention.
The objective of channel-wise attention is refine cross-channel embeddings by directly modelling the correlations between the different channels.
Hu et al. proposed 
squeeze and excitation units, multi-layer perceptron (MLP) networks that generate attention maps using
a dimensionality reduced descriptor
in which cross-channel spatial information is aggregated.
The resulting attention maps are then used to recalibrate the channels of the convolutional feature \cite{hu2018squeeze}.
\begin{equation}
\begin{gathered}
\label{attention_channel_se}
\resizebox{0.625\hsize}{!}{
$\text{A$_c$} = \sigma(MLP(AvgPool(\text{F}))^{\mathbb{R} \in C\text{x}1\text{x}1})$ \hfill
}
\end{gathered}
\end{equation}
where $\text{A}_c$ is the resulting channel attention map. F represents an intermediate convolutional feature map of size C x H x W.
$AvgPool$ represents adaptive average pooling. MLP is a multi-layer perception network and $\sigma(\cdot)$ is a sigmoid activation function. Subsequently, a refined feature map {F$'$} is computed using element-wise multiplication such that $\text{F}' = \text{F} \otimes  \text{A}_c$.

Thereafter, Woo et al. proposed convolutional block attention module (CBAM), in which they extend the concept of squeeze and excitation by adding a subsequent spatial attention gate \cite{woo2018cbam}. 
Spatial attention utilises the inter-spatial relationship within the convolutional features
to assist the CNN 
learning \textit{where} to attend within a feature map.
This was achieved by projecting the
channels of the convolutional feature into a
2D map embedding using channel-wise pooling operators, the resulting map was then 
passed through a convolutional layer followed by a sigmoid function to generate the final spatial attention map.
Moreover, in contrast to squeeze and excitation units, \cite{woo2018cbam} demonstrate that incorporating max pooling in addition to average pooling can provide complementary clues that can enhance the overall attention performance.
Accordingly, cross-channel and spatial attention can be described as follows:
\begin{equation}
\begin{gathered}
\label{attention_channel_cbam}
\resizebox{0.92\hsize}{!}{
$\text{A$_c$} = \sigma(MLP(AvgPool(\text{F})) + MLP(MaxPool(\text{F})))^{\mathbb{R} \in C\text{x}1\text{x}1}$ \hfill
}
\end{gathered}
\end{equation}
\begin{equation}
\begin{gathered}
\label{attention_spatial_cbam}
\resizebox{0.925\hsize}{!}{
$\text{A$_s$} = \sigma(Conv2D_{3\text{x}3}(AvgPool(\text{F}) + MaxPool(\text{F})))^{\mathbb{R} \in C\text{x}H\text{x}W}$ \hfill
}
\end{gathered}
\end{equation}
where $\text{A}_c$ and $\text{A}_s$ are the channel and spatial attention maps, respectively. F represents an intermediate convolutional feature map of size C x H x W.
$AvgPool$ and $MaxPool$ are adaptive average and max pooling layers. MLP is a multi-layer perception network. $Conv2D_{3\text{x}3}$ is a convolutional layer with kernel size of 3 x 3. $\sigma(\cdot)$ is a sigmoid activation function. 
Note that in the case of spatial attention, pooling operations are performed along the channel axis. 
Consequently, a refined feature map {F$'$} is computed using element-wise multiplication of the intermediate feature map F and the resulting attention map M, where M $\in \{\text{A}_c, \text{A}_s\}$.
Moreover, \cite{woo2018cbam} shows that combining both 
% their 
% channel and spatial
attention
approaches
% attention
by applying them 
in a consecutive order (i.e. channel-wise followed by spatial attention) can further enhance the overall performance in contrast to using either of the attention approaches individually, or using other combination schemes (e.g. spatial followed by channel-wise attention, or by applying both approaches in parallel).
Incorporating channel and spatial attention in CBAM can improve the representational capabilities of CNNs, and therefore enhances their performance \cite{woo2018cbam}.
This is in line with the findings of
Sun et al. \cite{sun2021attention} (pulmonary nodule classification in CT images),
Lu et al. \cite{lu2021multi} (pulmonary tuberculosis detection in CT images),
Nawshad et al. \cite{nawshad2021attention} (COVID-19 detection in X-ray images),
Sangeroki et al. \cite{sangeroki2021fast} (thoracic disease detection in chest X-rays),
and Park et al. \cite{park2018bam} (generic object detection).
In this work, we evaluate channel and spatial attention mechanisms for both the candidate proposal and the false positive reduction stage. We also present two fully convolutional attention 
blocks 
based on channel and spatial attention and demonstrate their effectiveness for the pulmonary nodule detection task in contrast to state-of-the-art attention approaches.

Global attention on the other hand, aims to model long-range correlations and inter-dependencies between arbitrary positions. 
Vaswani et al. proposed Transformer networks based on multi-headed self-attention for sequence-to-sequence tasks. An input vector is used in three ways, \textit{query}, \textit{key}, and \textit{value}.
Accordingly, attention is expressed
as a mapping between a \textit{query} and a set of \textit{key} and \textit{value} pairs, to an output vector, by finding the weighted sum of the \textit{values} using weights that are found by a compatibility function of the query and the corresponding key \cite{vaswani2017attention}. 
More formally, the over all attention process is described as follows:
\begin{equation}
\begin{gathered}
\label{attention_transformer}
\text{MultiHead}(Q,K,V) = \text{Concat}(head_1,\resizebox{0.02\hsize}{!}{..},head_i)W^O \hfill \\
\text{head$_i$} = \text{Attention}(QW^Q_i, KW^K_i, VW^V_i) \hfill \\
\text{Attention}(Q,K,V) = \text{softmax}(\frac{QK^T}{\sqrt{d_k}})V \hfill \\
\end{gathered}
\end{equation}
where $Q$, $K$, and $V$ represent \textit{queries}, \textit{keys}, and \textit{values}, and their correspondent learnable parameters $W^Q$, $W^K$, and $W^Q$, respectively. $W^O$ represents a learnable linear projection process.

Dosovitskiy et al. generalised this concept to computer vision tasks by splitting an image into a sequence of vectorised patches that can then be managed by a pure Transformer \cite{dosovitskiy2020image}.
While Transformers demonstrate a great potential in computer vision tasks, they rely on heavy pre-training and are difficult to scale to large inputs due to their computational cost. \cite{dosovitskiy2020image,wang2021transbts,chen2021transunet}. Moreover, vision transformers suffer when modeling local structures due to the tokenisation of input images \cite{yuan2021tokens}.
To tackle these limitations, a number of studies proposed using a hybrid architecture that combines both, CNNs for their spatial representation power and their relatively low computational cost, along with Transformers for their ability in modeling long-range dependencies.
A CNN encoder was used to extract embeddings from 3D MRI (magnetic resonance imaging) images, the down-sampled features were then fed into a pure Transformer unit in which output
was passed into a decoder CNN to perform object segmentation \cite{wang2021transbts}.
Other works used a similar approach to perform segmentation in 2D images \cite{chen2021transunet,zheng2021rethinking}.
In this work, due to the recent success and the increasing interest in vision transformers, we evaluate Transformer \cite{wang2021transbts} in the pulmonary nodule detection task and compare it against our proposed attention approach.

\subsection{Pulmonary nodule detection}

The increasing amounts of pulmonary data collections along with the advances in convolutional neural networks have attracted research interest to the pulmonary nodule localisation and classification problem.
Generally, most existing works deploy two stage CNN based detectors and investigate different levels of feature dimensionality, e.g, 2D, pseudo 3D (cross-sectional 2D planes), and 3D \cite{riquelme2020deep,dataset:LUNA16}.
Berens et al. \cite{berens2016znet} proposed a two stage detector based on a 2D U-Net \cite{ronneberger2015u} for region proposal, followed by a false alarm reduction CNN that takes three orthogonal 2D slices as an input. 
They evaluate their approach on LUNA16 dataset (\cite{dataset:LUNA16}), and conclude that incorporating 3D information may be a good direction to improve on the performance of their 2D approach.
Indeed, this was demonstrated in \cite{shi2018lung}, where the authors compare 2D, pseudo 3D, and 3D CNNs for the lung nodule detection task and find that using 3D kernels significantly improves the performance.
Similarly, Riquelme et al. demonstrate in their extensive survey on DL for lung nodule detection \cite{riquelme2020deep}, that the best performing methods are the ones that incorporate 3D information in their approaches.
This shows that 2D based approaches fail to fully exploit the
inherent 
3D nature of the nodule structure.

Liao et al. proposed a 3D region proposal network based on Faster RCNN \cite{ren2015faster} for the candidate proposal stage. The top five suspicious proposals are passed into a subsequent CNN in which a modified \textit{or-gate} \cite{pearl2014probabilistic} is employed to predict the final score \cite{liao2019evaluate}.
Zhu et al. proposed a similar 3D CNN based on dual path networks (DPN) \cite{NIPS2017_dualpathnetworks}
for nodule detection along with a gradient boosting machine (GBM) \cite{friedman2001greedyGBM} for the final classification stage \cite{method:deeplung}. They deploy an encoder-decoder design for their detection network to allow learning nodule features from different semantic levels.
They show that by using grouped convolutions \cite{method:resnext} and dense convolutional connections \cite{huang2017densely} in their DPN, the network was able to detect more nodules while decreasing the computational overhead.

Building on \cite{liao2019evaluate} and \cite{method:deeplung},
% Li et al. 
\cite{method:deepseed} adopted squeeze and excitation paths (\cite{hu2018squeeze}) to their detection CNN
to assist the network learn inter-dependencies within the extracted features .
This encoder-decoder design was also exploited by
% \cite{tang2019end},
\cite{chen2020end} and \cite{tang2018automated}, and have demonstrated great potential within the lung nodule detection task. Hence, we adopt this 3D encoder-decoder structure for our candidate proposal stage.
Additionally, both \cite{chen2020end} and \cite{method:deepseed} adopt focal loss \cite{lin2017focal} to tackle
the class imbalance in the pulmonary images and have empirically demonstrated an enhanced performance in the nodule detection task. We therefore adopt focal loss in our framework.

Moreover, both 3DDUALRPN \cite{Xie2017} and APNDCTDCNN \cite{method:accurateluna} used RPNs as the foundation of their methods. \cite{method:accurateluna} incorporated a 2D Faster RCNN detector to propose candidates from axial slices in the first stage, and employed a 3D CNN for the false positive reduction. While 3DDUALRPN \cite{Xie2017} proposed a dual path network consisting two U-shaped sub-networks, ResNet50 based RPN and DenseNet based detector, candidates from the two networks are combined using a label fusion method based on non maximum suppression (NMS) to produce the final detection. Both methods demonstrate excellent results on the LUNA16 dataset; however, they are relatively complex and require substantial computational resources as we demonstrate in Section \ref{section:experiments}.

For the false positive reduction stage, \cite{polat2018false} proposed using an ensemble of five CNNs to analyse nodules using different levels of context.
Each CNN was trained using a unique crop size (i.e. level of context) and the final score was defined as a function of all the predictions. In the same line, \cite{sun2021attention} proposed a more efficient approach by jointly analysing nodules on different contextual levels using a single CNN. Both experiments demonstrate that incorporating contextual information leads to an enhanced prediction. This is expected since pulmonary nodules have highly variable morphology and size (see Fig. \ref{fig:nodule_size_distribution_luna16}). Thus, we employ a joint analysis approach in our false positive reduction stage that aggregates multi contextual level features.

\begin{figure*}[t]
\centering

\includegraphics[width=1.0\linewidth]{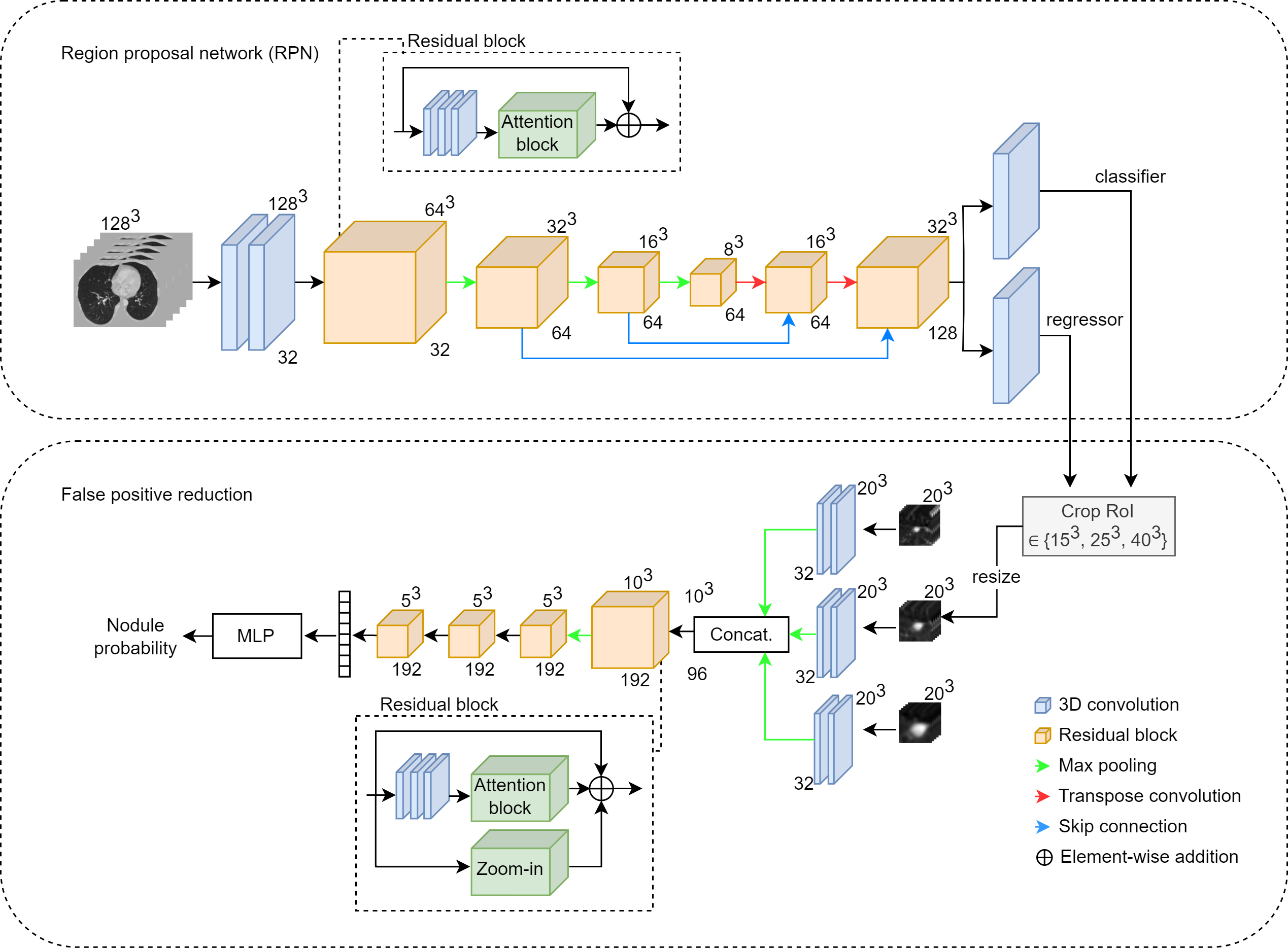}

\caption{The framework of AttentNet. AttentNet performs pulmonary nodule detection in two stages, candidate proposal, in which we exploit a 3D encoder-decoder network to predict suspicious nodule locations, and a false positive reduction stage in which a 3D CNN is used to extract deep features from the proposed nodules and produce the final prediction. We augment the building blocks of our network with attention units to assist the network in focusing on effective nodule features and therefore produce a more robust detections.
}

\label{fig:attentnetframework}
\end{figure*}

\section{Methodology}
\label{section:methodology}

Our proposed framework, AttentNet, detects pulmonary nodules in two stages: 1) candidate proposal, in which we adopt a 3D encoder-decoder CNN 
to propose suspicious locations of nodules at high sensitivity, 2) and a subsequent false positive reduction stage to reduce the number of false alarms. For the candidate proposal stage, we incorporate fully convolutional attention blocks to efficiently assist the network in focusing on informative features along both, inter-channel and spatial axes.
For the false positive reduction stage, we exploit nodule morphology by jointly analysing inputs of different contextual levels. Additionally, we use a zoom-in convolutional paths to allow the network pick fine details from different feature scales.
During inference time,
detections are aggregated from different
transformations of the input image.
The final classification score is found by ensembling the models of both stages.
The two detection stages are optimised separately, each according to its correspondent objective.
An overview of our proposed framework is presented in fig. \ref{fig:attentnetframework}.

\subsection{Candidate proposal stage}
\label{section:candidateproposalstage}

We incorporate an encoder-decoder network as the backbone of our candidate proposal stage.
An input image is first processed by two consecutive 3D convolutional layers of 32 channels each, using 3 x 3 x 3 kernels, followed by a 3D max pooling layer to reduce the spatial dimension of the resulting feature map by a factor of 2.
The resulting volumetric feature is then passed into a sequence of four residual blocks \cite{method:resnet} that consist of 2, 3, 3 and 3 residual units, respectively. 
Each of the residual blocks is followed by a max pooling layer. The resulting embeddings are then up-sampled by two subsequent deconvolutional layers using 2 x 2 x 2 kernels and strides of 2. Each up-sampling layer is followed by a residual block (consisting of 3 residual units each).
To avoid overfitting, we regularise the network using 3 dropout layers in the encoder part and one dropout layer in the last decoder layer. We use 0.3 dropout rate for all dropout layers.

It is worth noting that at an effective image resolution, even a single image can exhaust the GPU memory. To address this constraint and allow processing the 3D inputs at an effective resolution and a representative training batch size, we utilise grouped convolutions \cite{method:resnext} within the residual units.
Grouped convolution allows efficient model parallelism by splitting both, the feature channels, and the convolutional kernels into a number groups. Each group of filters processes their correspondent feature channels and produce a part of the output channels.
The resulting feature maps are then projected into a new linear space using a subsequent 1 x 1 x 1 convolution.
This allows reducing the computational overhead and the number of parameters while retaining the number of features and a consistent performance 
% \cite{kopuklu2019resource,hara2018can,method:resnext}
\cite{kopuklu2019resource,method:resnext}.
Grouped convolution have repeatedly demonstrated effectiveness within the lung nodule detection task 
% \cite{method:deeplung,zhang2019classification,gu40retrieval,zhang2020classification}.
\cite{method:deeplung,zhang2019classification,gu40retrieval}.
Accordingly, a single residual unit consists of three convolutional layers of 1 x 1 x 1, 3 x 3 x 3 , and 1 x 1 x 1 kernel sizes, respectively, with the second layer utilising grouped convolutions of 32 groups
% \cite{kopuklu2019resource,hara2018can,method:resnext}
\cite{kopuklu2019resource,method:resnext}.
An overview of our candidate proposal backbone network is presented in Fig. \ref{fig:attentnetframework}.

Inspired by the success demonstrated by attention mechanisms in medical imaging related tasks (e.g.
% \cite{fu2021multimodal,xiao2021pam,sun2021attention,lu2021multi,nawshad2021attention,sangeroki2021fast})
\cite{sun2021attention,lu2021multi,nawshad2021attention,sangeroki2021fast})
, as well as the sparsity of pulmonary nodule locations, and the high relevance of nodule morphology for the nodule detection task, we utilise
two attention mechanisms within the building blocks of our backbone network to assist the network in focusing at meaningful and effective embeddings.
Specifically, we propose 3D fully convolutional cross-channel and spatial attention blocks for the candidate proposal stage.

\begin{figure}[t]
\centering

\includegraphics[width=1.0\linewidth]{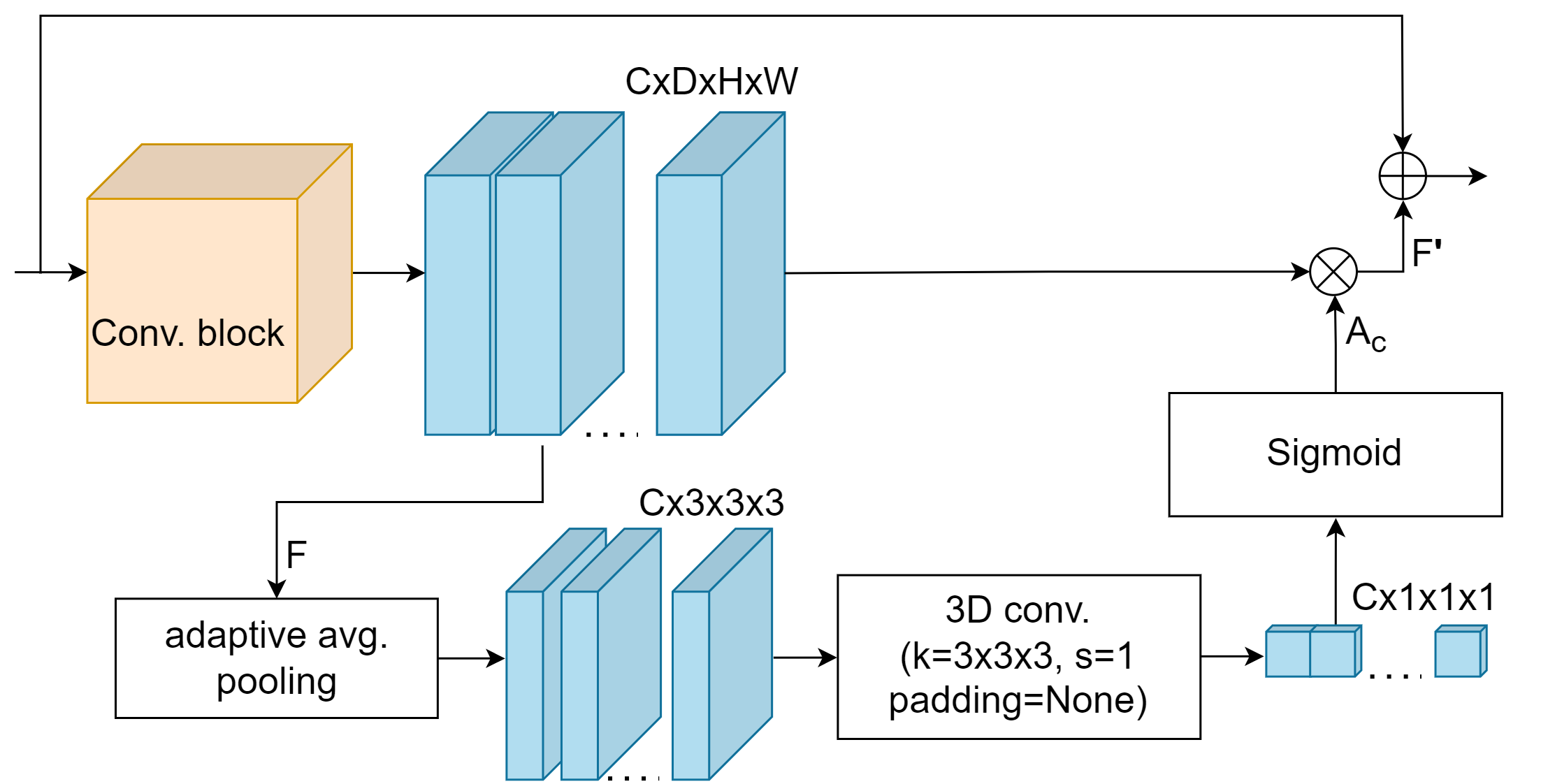}

\caption{
An overview of our proposed 3D fully convolutional cross-channel attention unit within a residual block. As illustrated, our channel attention exploits 3D adaptive pooling to embed spatial information from an intermediate convolutional feature map $\text{F}$, these are passed into a 3D convolutional layer in which output is an attention map $\text{A}_c$ of size ${C\text{~x~}1\text{~x~}1\text{~x~}1}$. This will then be used to adaptively refine the intermediate feature maps inferring channel importance and inter-channel correlations ($\text{F}'$). Here, $\bigotimes$ and $\bigoplus$ represent element-wise multiplication and addition, respectively.
% \textcolor{violet}{
Note that the addition operation represents the residual path in the residual block.
% }
The parameter k represents the kernel size used in the convolutional layers.
}

\label{fig:channelattentionblock}
\end{figure}

Given an intermediate 3D convolutional feature map $\text{F} \in \mathbb{R}^{C\text{x}D_F\text{x}H_F\text{x}W_F}$,
cross channel attention is computed as follows:
First, spatial information is aggregated using adaptive average pooling, resulting an embedding 
$\text{E} \in \mathbb{R}^{C\text{x}D_E\text{x}H_E\text{x}W_E}$.
Here, ${C}$ represents the channels of the feature map, ${D_F\text{,~}H_F\text{,~}W_F}$ and ${D_E\text{,~}H_E\text{,~}W_E}$ represent the depth, height, and width of F and E, respectively. 
Note that in our preliminary experiments, we test other types of spatial feature aggregation (e.g. adaptive max pooling) and observe no particular benefit within our nodule detection task, therefore we continue using average pooling.
Consequently, E is passed into a convolutional layer with 3D kernels of size ${D_E\text{~x~}H_E\text{~x~}W_E}$, and no padding, producing a ${C\text{~x~}1\text{~x~}1\text{~x~}1}$ descriptor. A sigmoid function is then applied to produce the final channel attention map $\text{A$_{c}$}$, in which channel-wise importance scores are predicted. The feature map F is then recalibrated using element-wise multiplication with the attention map A. 
For our experiment, we set ${D_E\text{,~}H_E\text{,~}W_E}$ to 3, therefore, the overall attention map computation can be summarised as follows:
\begin{equation}
\begin{gathered}
\label{attention_channel_1}
\text{A$_{c}$} = \sigma(\text{W}(\text{E}))^{\mathbb{R} \in C\text{x}1\text{x}1\text{x}1} \hfill \\
~~~~ = \sigma(Conv3D_{3\text{x}3\text{x}3}(AvgPool3D(\text{F})^{\mathbb{R} \in C\text{x}3\text{x}3\text{x}3})) \hfill
\end{gathered}
\end{equation}
where $AvgPool3D$ represents adaptive 3D average pooling. $Conv3D_{3\text{x}3\text{x}3}$ is a 3D convolutional layer, the subscript represents the size of the kernels used in the convolutional layer.
$\sigma(\cdot)$ is a sigmoid activation function. Note that we continue using these notations for the reminder of the article. Accordingly, the refined feature map is computed as follows:
\begin{equation}
\begin{gathered}
\label{attention_channel_2}
\text{F$'$} = \text{A$_{c}$} \otimes \text{F} \hfill 
\end{gathered}
\end{equation}
where $\otimes$ denotes element-wise multiplication. Our cross-channel attention block is visualised in Fig. \ref{fig:channelattentionblock}

Our channel-wise attention strategy aims to assist the network in effectively learning to focus on informative feature and ignore (down-weight) irrelevant or less informative ones. Other works that incorporate multi-layer perceptrons rely on heavy dimensionality reduction to decrease the computation overhead of the attention network. Particularly, \cite{hu2018squeeze,woo2018cbam} use an adaptive pooling operation to create a ${C\text{~x~}1\text{~x~}1\text{~x~}1}$ descriptor in which spatial information of the
% 2D
intermediate features is embedded. This is then passed to a multi-layer perceptron where the feature's dimensionality is further reduced by a pre-defined factor (see Eqs. \ref{attention_channel_se} and \ref{attention_channel_cbam}).
Note that the size of the spatial descriptor is less proportional to the intermediate embedding when dealing with 3D feature maps (e.g. pulmonary nodules) in contrast to that in 2D based analysis (e.g. \cite{hu2018squeeze,woo2018cbam}),
leading to a limited spatial information.
We avoid this by replacing the multi-layer perceptron by a fully convolutional network, allowing 
an efficient use of rich spatial descriptors with higher dimensionality (i.e. ${C\text{~x~}3\text{~x~}3\text{~x~}3}$).

\begin{figure*}[t]
\centering

\includegraphics[width=1.0\linewidth]{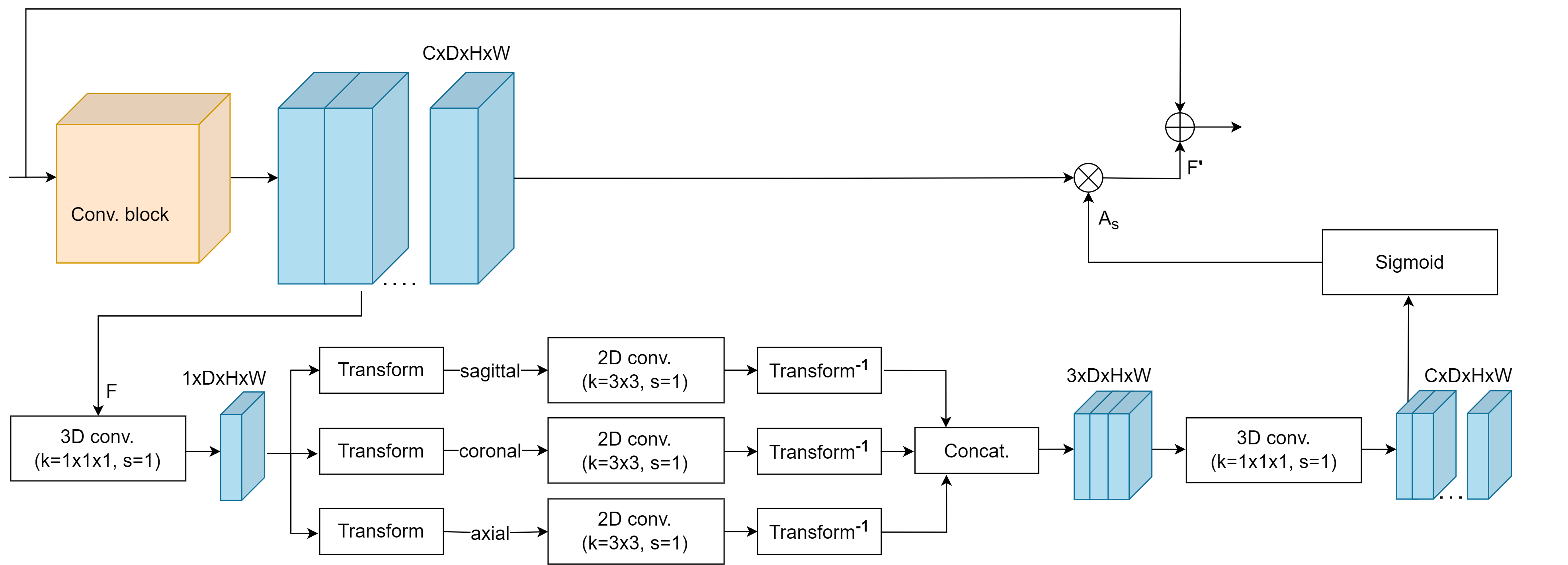}

\caption{
An overview of our proposed 3D fully convolutional inter-spatial attention unit within a residual block.
Our spatial attention takes as input an intermediate 3D feature map $\text{F}$ of ${C}$ channels, projects it into a ${1}$ channel feature map (using ${1\text{~x~}1\text{~x~}1}$ convolutions) that is then transformed into three orthogonal planes (axial, coronal, and sagittal). 
Each of the resulting features is then processed by a unique 2D convolutional layer to learn cross-sectional spatial representations.
% 
% These are then Transformed back into the original image plane, aggregated by concatenation, and are projected back to ${C}$ 3D channels in which we use to infer cross-sectional spatial attention.
% \textcolor{violet}{
The resulting feature maps are then spatially aligned, aggregated by concatenation, and are linearly projected back to ${C}$ 3D channels in which we use to infer cross-sectional spatial attention $\text{A}_s$.
% }
% 
Intermediate feature maps are adaptively refined ($\text{F}'$) using element-wise multiplication ($\bigotimes$ in the figure).
Here, $\bigoplus$ represent element-wise addition used in the residual path of the residual block. The parameters k and s represent the kernel size and the stride used in the convolutional layers.
}

\label{fig:spatialattentionblock}
\end{figure*}

To perform spatial attention, we adopt a joint analysis approach that integrates spatial information from different image cross-sections (i.e. axial, coronal, and sagittal).
% 
% \sout{For simplicity, we present our approach with an axial input
% % images
% , however, the same principle is applied into inputs from other cross-sectional planes (i.e. coronal and sagittal).}
% \textcolor{violet}{
% Given a \textcolor{violet}{3D} input image
% \sout{in the axial plane,}
First,
an intermediate 3D feature map $\text{F} \in \mathbb{R}^{C\text{x}D_F\text{x}H_F\text{x}W_F}$ is linearly transformed into $\text{E} \in \mathbb{R}^{1\text{x}D_F\text{x}H_F\text{x}W_F}$ using a convolutional layer with kernels of size ${1\text{~x~}1\text{~x~}1}$:
\begin{equation}
\begin{gathered}
\label{attention_spatial_1}
\resizebox{0.925\hsize}{!}{
$\text{E} = Conv3D_{1\text{x}1\text{x}1}(F^{\mathbb{R} \in C\text{x}D_F\text{x}H_F\text{x}W_F})^{\mathbb{R} \in 1\text{x}D_F\text{x}H_F\text{x}W_F} \hfill $
}
\end{gathered}
\end{equation}
Note that the resulting feature map $\text{E}$ is of dimensions ${1\text{~x~}D_F\text{~x~}H_F\text{~x~}W_F}$, therefore, by reducing the first axis (i.e. ${D_F\text{~x~}H_F\text{~x~}W_F}$), $\text{E}$ can be processed using 2D convolutions along the $D_F$ axis (this is equivalent to having a 2D input of $D_F$ channels).
The feature map
$\text{E}$
is now projected into the axial, coronal and sagittal planes.
% \sout{resulting a total of three feature projections (axial, coronal, and sagittal).}
Each of these feature maps
is then processed by a unique 2D convolutional network, producing three 2D convolutional feature maps~$\{\text{E}_{axi},\text{E}_{cor},\text{E}_{sag}\} \in \mathbb{R}^{D_F\text{x}H_F\text{x}W_F}$:
% \begin{equation}
% \begin{gathered}
% \label{attention_spatial_2}
% \text{E$_{axi}$} = Conv2D_{3\text{x}3}(\text{E})^{\mathbb{R} \in D_F\text{x}H_F\text{x}W_F} \hfill \\
% \text{E$_{cor}$} = Conv2D_{3\text{x}3}(Coronal(\text{E}))^{\mathbb{R} \in D_F\text{x}H_F\text{x}W_F} \hfill \\
% \text{E$_{sag}$} = Conv2D_{3\text{x}3}(Sagittal(\text{E}))^{\mathbb{R} \in D_F\text{x}H_F\text{x}W_F} \hfill
% \end{gathered}
% \end{equation}
% 
% 
\begin{equation}
\begin{gathered}
\label{attention_spatial_2}
\text{E$_{axi}$} = Conv2D_{3\text{x}3}(Axial(\text{E}))^{\mathbb{R} \in D_F\text{x}H_F\text{x}W_F} \hfill \\
\text{E$_{cor}$} = Conv2D_{3\text{x}3}(Coronal(\text{E}))^{\mathbb{R} \in D_F\text{x}H_F\text{x}W_F} \hfill \\
\text{E$_{sag}$} = Conv2D_{3\text{x}3}(Sagittal(\text{E}))^{\mathbb{R} \in D_F\text{x}H_F\text{x}W_F} \hfill
\end{gathered}
\end{equation}
where $Axial(\cdot)$ $Coronal(\cdot)$ and $Sagittal(\cdot)$ are transformation functions that project
inputs from axial plane into coronal and sagittal planes, respectively. 
% 
% 
% \textcolor{violet}{
This is equivalent to applying 2D convolution with different absolute orientations in the 3D space.
% }
% 
% 
% The resulting embeddings are then projected back to the original plane (axial), and are combined (concatenated) along a new axis to form a cross-sectional (3D) feature map $\text{E}_{cs} \in \mathbb{R}^{3\text{x}D_F\text{x}H_F\text{x}W_F}$:
% \textcolor{violet}{
The resulting embeddings are then spatially aligned, and are combined (concatenated) along a new axis to form a cross-sectional (3D) feature map $\text{E}_{cs} \in \mathbb{R}^{3\text{x}D_F\text{x}H_F\text{x}W_F}$:
% }
% \begin{equation}
% \begin{gathered}
% \label{attention_spatial_3}
% \resizebox{0.925\hsize}{!}{
% $\text{E$_{cs}$} = Concat(\text{E$_{axi}$}, Axial(\text{E$_{cor}$}), Axial(\text{E$_{sag}$}))^{\mathbb{R} \in 3\text{x}D_F\text{x}H_F\text{x}W_F}$ \hfill
% }
% \end{gathered}
% \end{equation}
% 
% 
\begin{equation}
\begin{gathered}
\label{attention_spatial_3}
\resizebox{0.88\hsize}{!}{
$\text{E$_{cs}$} = Concat(\text{E$_{axi}$}, \text{E$_{cor}$}, \text{E$_{sag}$})^{\mathbb{R} \in 3\text{x}D_F\text{x}H_F\text{x}W_F}$ \hfill
}
\end{gathered}
\end{equation}
Last, $\text{E}_{cs}$ is transformed back into $\mathbb{R}^{C\text{x}D_F\text{x}H_F\text{x}W_F}$ using ${1\text{~x~}1\text{~x~}1}$ convolution. A sigmoid function is then applied to produce the final spatial attention map $\text{A$_{s}$}$, in which spatial importance scores are predicted:
\begin{equation}
\begin{gathered}
\label{attention_spatial_4}
\text{A$_{s}$} = \sigma(Conv3D_{1\text{x}1\text{x}1}(\text{E$_{cs}$})^{\mathbb{R} \in C_F\text{x}D_F\text{x}H\text{x}W_F}) \hfill
\end{gathered}
\end{equation}
The attention map $\text{A$_{s}$}$ is used to refine the intermediate feature map using element-wise multiplication:
\begin{equation}
\begin{gathered}
\label{attention_spatial_5}
\text{F$'$} = \text{A$_{s}$} \otimes \text{F} \hfill \\
\end{gathered}
\end{equation}
The overall proposed spatial attention process is visualised in Fig. \ref{fig:spatialattentionblock}

% channel attention :

% spatial attention :

While spatial attention proposed in \cite{woo2018cbam} focuses on capturing spatial correlations within the convolutional features, we argue that
the use of element-wise pooling operations along the channel axis, to aggregate spatial information 
limits the volumentric information when dealing with 3D data. Note that spatial attention in \cite{woo2018cbam} was originally designed for 2D based analysis,
% \textcolor{violet}{
see Eq. \ref{attention_spatial_cbam}.
% }

\begin{figure}[t]
\centering
\begin{tabular}{cccccc}

\includegraphics[width=0.148\linewidth]{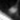} 
\includegraphics[width=0.148\linewidth]{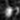} 
\includegraphics[width=0.148\linewidth]{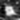} 
\includegraphics[width=0.148\linewidth]{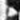} 
\includegraphics[width=0.148\linewidth]{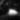} 
\includegraphics[width=0.148\linewidth]{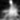} \\

\includegraphics[width=0.148\linewidth]{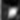} 
\includegraphics[width=0.148\linewidth]{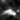} 
\includegraphics[width=0.148\linewidth]{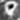} 
\includegraphics[width=0.148\linewidth]{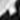} 
\includegraphics[width=0.148\linewidth]{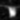} 
\includegraphics[width=0.148\linewidth]{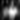} \\

\includegraphics[width=0.148\linewidth]{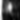} 
\includegraphics[width=0.148\linewidth]{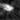} 
\includegraphics[width=0.148\linewidth]{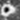} 
\includegraphics[width=0.148\linewidth]{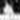} 
\includegraphics[width=0.148\linewidth]{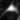} 
\includegraphics[width=0.148\linewidth]{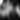} \\
% axial ~~~~~~~~~~~~
% coronal ~~~~~~~~~~~~
% sagittal \\

\end{tabular}

\caption{Pulmonary nodules viewed in different cross-sectional planes: axial (top), coronal (middle), and sagittal (bottom).
}
\label{fig:nodule_cross_sections}
\end{figure}

%
Our cross-sectional spatial attention approach not only leverages 2D inter-spatial relations, but also captures 3D information by jointly analysing the three orthogonal planes of the 3D feature map (axial, coronal, and sagittal). This particularly important when managing volumetric images, where target structures may have different visual appearance when viewed in different cross-sectional planes, e.g., pulmonary nodules. See Fig. \ref{fig:nodule_cross_sections}.

Our attention blocks (cross-channel, or spatial attention) can be straightforwardly integrated and jointly trained with any 3D CNN architecture. In line with \cite{woo2018cbam} and \cite{hu2018squeeze}, we place our proposed attention block prior to the residual path of a residual unit.
In our experiment, we individually evaluate the benefits of both, channel and spatial attention, within the pulmonary nodule detection task.
We further evaluate the impact of incorporating both attention techniques in combination.
More details are presented in Section \ref{section:experiments}.

To perform the final detection, the output of the feature extraction network is passed into two parallel convolutional layers, a classification and a regression layer, to predict classes and nodule coordinates for each position in the feature map, respectively.
In line with \cite{tang2018automated,method:deeplung,method:deepseed}, we train our network with 3 reference anchor sizes, 5, 10, and 20, set based on the nodule size distribution (see Fig. \ref{fig:nodule_size_distribution_luna16}).
An anchor is considered to be positive if it has an intersection over union (IoU) $\geq$ 0.5, and negative if IoU $<$ 0.2. 
The network is trained according to the combined loss function:
\begin{equation}
\begin{gathered}
\label{detection_overall_loss}
L = 
\lambda~L_{cls} 
~+~
p^\ast L_{reg}
\end{gathered}
\end{equation}
where $L_{cls}$ and $L_{reg}$ are the classification and location regression losses, respectively. $\lambda$ is a balancing operator and is set to 1 in our experiment. $p^\ast \in \{1,0\}$ denoting positive and negative anchors, respectively. Similar to \cite{method:deepseed,chen2020end}, we adopt a focal cross-entropy loss \cite{lin2017focal} for the nodule classification task:
\begin{equation}
\begin{gathered}
\label{focal_loss}
L_{cls}(p_t) = -\alpha (1-p_t)^{\gamma}~\text{log}(p_t)
\end{gathered}
\end{equation}
here, considering an anchor's predicted probability $p$, $p_t=p$ if the ground-truth label is positive, and $p_t=(1-p)$ otherwise. $\gamma$ is a modulating parameter. Well classified samples (i.e. $p_t \rightarrow 1$) cause the modulating term to approach 0, down-weighting their loss values, and vice versa for hard examples. Note that $\alpha$ acts as a class balancing parameter. We find that setting $\gamma$ and $\alpha$ to 2 and 0.5, respectively, provides a favorable balanced performance.
In line with the essence of attention, focal loss assists the network in focusing on more informative samples and therefore, structures. This is particularly important when managing class imbalanced data, in this case, pulmonary nodule images.

Moreover, we use smooth L1 loss \cite{ren2015faster} for location regression task :
\begin{equation}
\begin{gathered}
\label{smooth_l1_loss}
L_{reg}(t^\ast, t) = \begin{cases}
\left | t^\ast-t \right | & \text{~if~} 
\left |t^\ast-t\right |> 1 \\ 
(t^\ast-t)^2 & \text{otherwise}
\end{cases}
\end{gathered}
\end{equation}
given that $t$ and $t^\ast$ are vectors representing the relative coordinates of a nodule location in, respectively, the prediction space and ground-truth:
\begin{equation}
\begin{gathered}
\label{predicted_relative_coords}
t_i = (\frac{x-x_a}{d_a}, \frac{y-y_a}{d_a}, \frac{z-z_a}{d_a}, \text{log}\frac{d}{d_a})
\end{gathered}
\end{equation}
where $(x, y, z, d)$ and $(x_a, y_a, z_a, d_a)$ are the predicted nodule center and diameter, and the coordinates of an anchor i, respectively.
Similarly, the ground-truth relative coordinates are defined as $t^{\ast}_i$ using the original nodule coordinates $(x^{\ast}, y^{\ast}, z^{\ast}, d^{\ast})$.

To further reduce the class imbalance impact, we adopt an online hard negative mining strategy \cite{method:OHEM}.
During training, $N$ negative samples with the highest nodule probability are selected to be used in the computation of the loss.
Remaining samples are ignored and do not contribute to the loss. Here, we set $N$ to 2.
Hard negative mining has repeatedly demonstrated 
usefulness in the nodule detection task 
% \cite{tang2018automated,liao2019evaluate,method:deeplung,method:deepseed,chen2020end,tang2019end}
\cite{tang2018automated,liao2019evaluate,method:deeplung,method:deepseed,chen2020end}.

\begin{figure}[t]
\centering
\begin{tabular}{c}

\frame{
\includegraphics[width=0.8\linewidth]{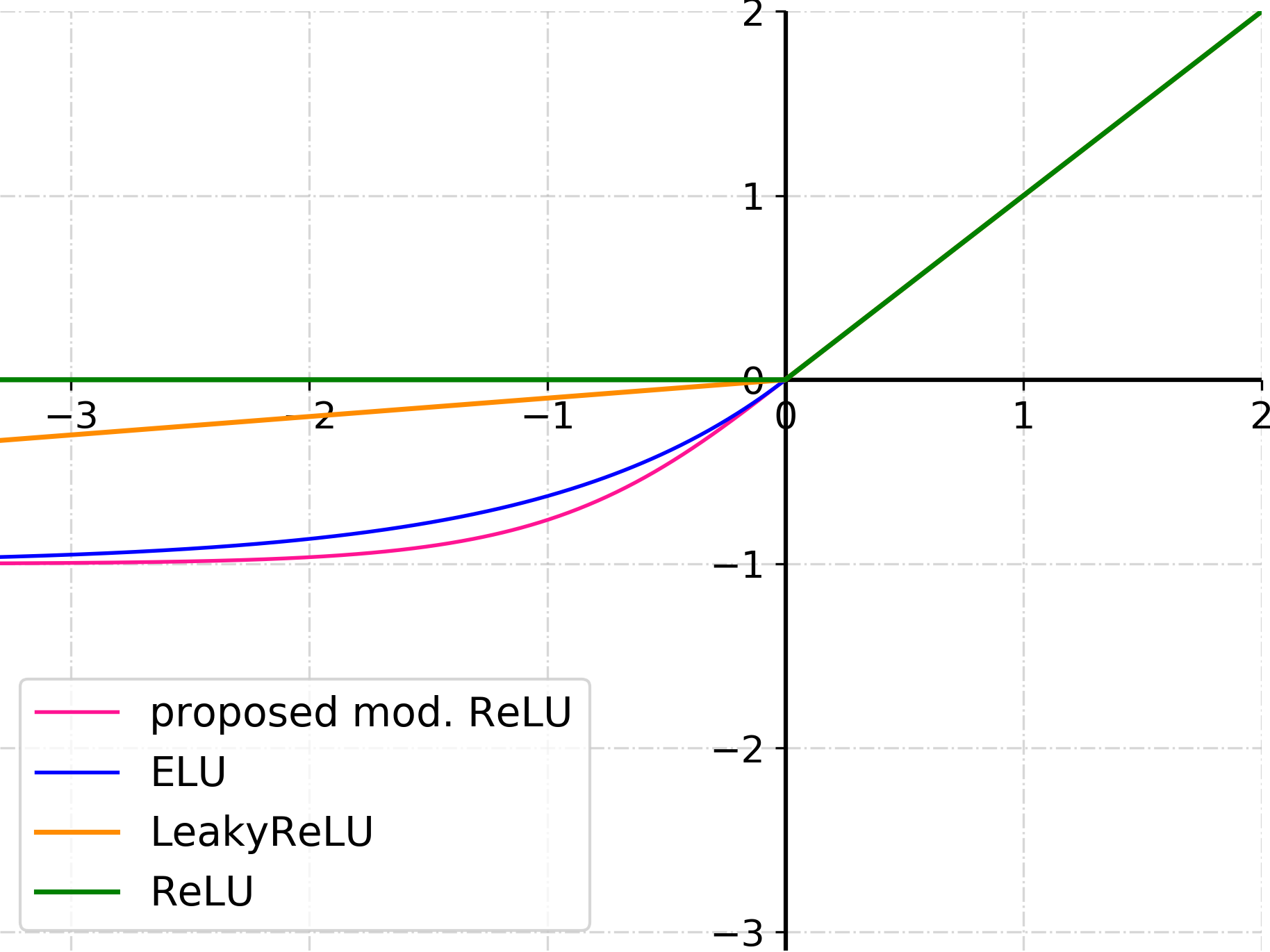}
} \\

\frame{
\includegraphics[width=0.8\linewidth]{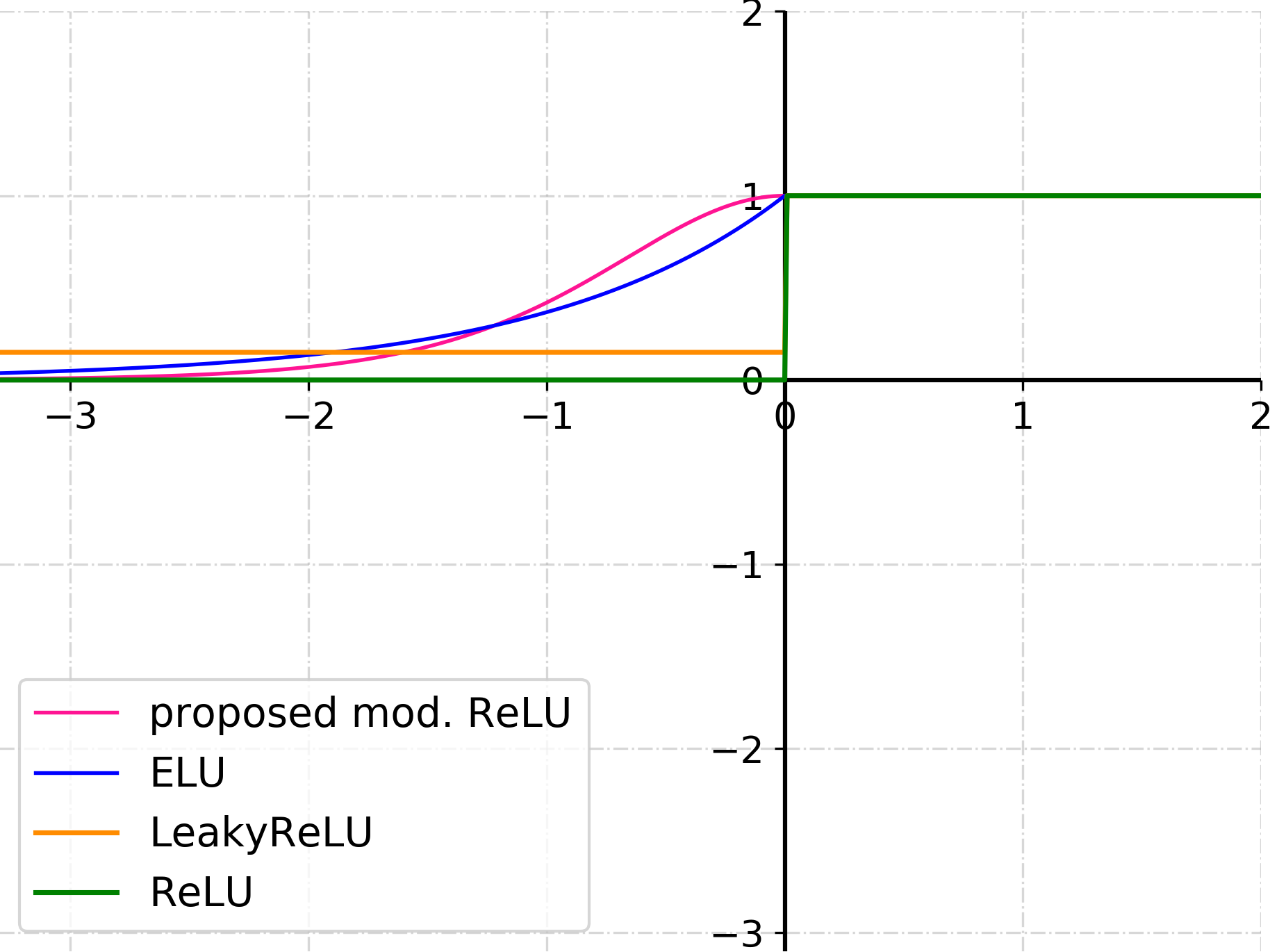}
} \\

\end{tabular}
\caption{Visual comparison of ReLU, Leaky ReLU, ELU, and our proposed modified ReLU activation (top), and their correspondent derivatives (bottom).}
\label{fig:activiation_functions}
\end{figure}

For the network's activation function, we use a modified rectified linear unit (ReLU). ReLU \cite{glorot2011deep} functions have gained popularity due to their simplicity and robustness.
This manifests in their ability of preserving the gradient flow in the positive input range. 
Since the linear portion of ReLU doesn’t saturate (i.e. unbounded), it allows gradients on active neurons to remain proportional to their activation.
Moreover, in the negative input range, ReLU promotes network sparsity by setting the activation value to zero \cite{glorot2011deep}.
However, this becomes a problem when many
inputs have negative values, leading to a degraded gradient flow in the backpropagation process, and therefore, no learning.
To avoid this problem, we simply modify ReLU such that it generates small negative outputs when the input lies in the negative range. Particularly, we incorporate a hyperbolic tangent function for the negative range of ReLU. Given a ReLU function $f(x) = max(0, x)$, our modified ReLU can be formulated as follows:
\begin{equation}
\begin{gathered}
\label{modified_relu}
g(x) =
\begin{cases}
x & \text{~if~} x>0  \\ 
tanh(x) & \text{otherwise}
\end{cases}
\end{gathered}
\end{equation}
where x is the input of the activation function. The idea of this modification is to preserve the linear characteristic of ReLU for positive input values, while improving the gradient flow for negative inputs. 
Moreover, unlike pure ReLU, the modified ReLU is smooth around the origin (x=0), promoting a faster learning process.
A visual comparison of both activations is presented in Fig. \ref{fig:activiation_functions}.
% 
% \sout{In Section \ref{section:experiments}, we empirically demonstrate the benefits of our proposed activation and compare it to pure ReLU activation.
% % 
% We also compare the performance of the proposed activation function against Leaky ReLU \cite{maas2013rectifier}, a commonly used ReLU alternative in which the negative gradient flow is enhanced by introducing a small slope for input values that lie within the negative range of ReLU.
% %
% While Leaky ReLU has been widely used within the pulmonary nodule detection task to reduce the risk of the dying neurons 
% % \cite{yu2020amplification,haibo2021improved,huang2019using,wang2019lung,tan2018fast}
% \cite{yu2020amplification,haibo2021improved,huang2019using}
% , we demonstrate in our experiment that our modified activation can outperform both, ReLU and Leaky ReLU in the nodule detection task.}
% 
In Section \ref{section:experiments}, we empirically demonstrate the benefits of our proposed activation and compare it to pure ReLU activation.
We also compare the performance of the proposed activation function against 
commonly used ReLU alternatives, Leaky ReLU \cite{maas2013rectifier} and ELU \cite{clevert2015fast}, in which the negative gradient flow is enhanced by allowing outputs for input values that lie within the negative range of ReLU.
It is worth noting that Leaky ReLU has been widely used within the pulmonary nodule detection task to reduce the risk of the dying neurons \cite{yu2020amplification,haibo2021improved,huang2019using,wang2019lung,tan2018fast}, we demonstrate in our experiment that our modified activation can outperform
ReLU, Leaky ReLU, and ELU activations in the nodule detection task.

Furthermore, we employ a testing time augmentation (TTA) strategy in which an input image is orientated along axial, coronal, and sagittal cross-sections. Each of the resulting images is then processed by the network to predict candidate locations.
Results from all cross-sections are then aggregated and used to form the final set of predictions
It is worth noting that these augmentations are similar to the ones used during training. Testing time augmentation is a simple and an effective way to enhance the performance of DNNs 
% \cite{wang2021transbts,moshkov2020test,simonyan2014very,szegedy2016,shanmugam2020and}
\cite{wang2021transbts,moshkov2020test,simonyan2014very}. 
We demonstrate this in our experiment for the nodule detection task.

% \begin{figure}[t]
% \centering
% \begin{tabular}{cc}

% ~~ ReLU ~~~~~~~~~~~~~~~~~~~~~
% Modified ReLU \\ 

% \fbox{\includegraphics[width=0.441\linewidth]{figures/activation_functions/relu.png}}
% \fbox{\includegraphics[width=0.441\linewidth]{figures/activation_functions/tanhleakyrelu.png}} \\

% % ~~ ReLU ~~~~~~~~~~~~~~~~~~~~~
% % Modified ReLU 

% \\

% \fbox{\includegraphics[width=0.441\linewidth]{figures/activation_functions/relu_derivative.png}}
% \fbox{\includegraphics[width=0.441\linewidth]{figures/activation_functions/tanhleakyrelu_derivative.png}} \\

% % ~~~~~~~~ReLU derivative~~~~~~~~
% % Modified ReLU derivative
% % \\

% \end{tabular}

% \caption{
% Visual comparison of ReLU and our modified activation (top), and their correspondent derivatives (bottom).}
% \label{fig:activiation_functions}
% \end{figure}

\subsection{False positive reduction stage}
\label{section:falsepositivereductionstage}

\begin{figure}[t]
\centering
\begin{tabular}{cccccc}

\includegraphics[width=0.148\linewidth]{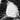}
\includegraphics[width=0.148\linewidth]{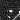}
\includegraphics[width=0.148\linewidth]{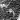}
\includegraphics[width=0.148\linewidth]{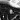}
\includegraphics[width=0.148\linewidth]{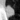}
\includegraphics[width=0.148\linewidth]{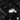} \\

\includegraphics[width=0.148\linewidth]{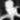}
\includegraphics[width=0.148\linewidth]{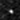}
\includegraphics[width=0.148\linewidth]{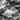}
\includegraphics[width=0.148\linewidth]{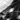}
\includegraphics[width=0.148\linewidth]{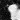}
\includegraphics[width=0.148\linewidth]{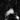} \\

\end{tabular}
\caption{
A visual comparison between pulmonary nodule (top) and non-nodule (bottom) regions. The two classes share similar appearance and morphology, increasing the complexity of the classification task.}
\label{fig:nodule_vs_non_nodules}
\end{figure}

Pulmonary nodules are highly variable in shape, size, and density, they have similar morphological characteristics to neighbouring organs and non-nodule structures, e.g., blood vessels and airways (see Figs \ref{fig:nodule_size_distribution_luna16} and \ref{fig:nodule_vs_non_nodules}).
This high morphological variability increases the complexity of the detection task leading to high rates of false positive detections
% \cite{dataset:LUNA16,larici2017lung,shen2021classification,zhao2003automatic,wozniak2018small,callister2015british,gould2007evaluation}
\cite{dataset:LUNA16,larici2017lung,shen2021classification,zhao2003automatic,wozniak2018small}.
To address this issue, we deploy a false positive reduction stage, in which we utilise a joint analysis based approach that incorporates nodule morphology and spatial context information to perform the 
% \textcolor{violet}{
final
% }
detection.
% detection.
% 
We further utilise cross-channel attention to assist the network in focusing on important information by modeling correlations between the embedded features. An overview of our false positive reduction stage is presented in Fig. \ref{fig:attentnetframework}.

% \begin{figure}[t]
% \centering
% \begin{tabular}{cc}
% \frame{
% \includegraphics[width=0.457\linewidth]{figures/activation_functions/compare_act_funcs.png} 
% }
% \frame{
% \includegraphics[width=0.457\linewidth]{figures/activation_functions/compare_act_funcs_dtv.png}
% } 
% \end{tabular}
% \caption{
% Visual comparison of ReLU, Leaky ReLU, and our proposed modified ReLU activation (left), and their correspondent derivatives (right).}
% \label{fig:activiation_functions}
% \end{figure}

% \begin{figure*}[t]
% \centering
% \begin{tabular}{cc}

% \frame{
% \includegraphics[width=0.45\linewidth]{figures/activation_functions/compare_act_funcs_4.0.png} 
% }
% \frame{
% \includegraphics[width=0.45\linewidth]{figures/activation_functions/compare_act_funcs_dtv_4.0.png}
% }

% \end{tabular}
% \caption{\textcolor{red}{[Figure Updated]} Visual comparison of ReLU, Leaky ReLU, ELU, and our proposed modified ReLU activation (left), and their correspondent derivatives (right).}
% \label{attentnetdetection_fig:activiation_functions}
% \end{figure*}

\begin{figure*}[t]
\centering
\begin{tabular}{cccccccccccc}
\includegraphics[width=0.0758\linewidth]{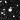}
\includegraphics[width=0.0758\linewidth]{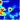} 
% ~
\includegraphics[width=0.0758\linewidth]{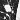}
\includegraphics[width=0.0758\linewidth]{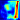} 
% ~
\includegraphics[width=0.0758\linewidth]{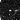}
\includegraphics[width=0.0758\linewidth]{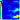} 
% ~
\includegraphics[width=0.0758\linewidth]{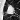}
\includegraphics[width=0.0758\linewidth]{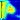} 
% ~
\includegraphics[width=0.0758\linewidth]{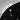}
\includegraphics[width=0.0758\linewidth]{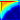} 
% ~
\includegraphics[width=0.0758\linewidth]{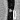}
\includegraphics[width=0.0758\linewidth]{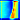} 
\\

\includegraphics[width=0.0758\linewidth]{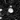}
\includegraphics[width=0.0758\linewidth]{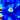}
% ~
\includegraphics[width=0.0758\linewidth]{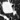}
\includegraphics[width=0.0758\linewidth]{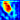} 
% ~
\includegraphics[width=0.0758\linewidth]{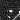}
\includegraphics[width=0.0758\linewidth]{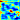} 
% ~
\includegraphics[width=0.0758\linewidth]{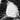}
\includegraphics[width=0.0758\linewidth]{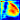} 
% ~
\includegraphics[width=0.0758\linewidth]{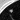}
\includegraphics[width=0.0758\linewidth]{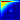} 
% ~
\includegraphics[width=0.0758\linewidth]{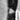}
\includegraphics[width=0.0758\linewidth]{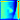} 
\\

\includegraphics[width=0.0758\linewidth]{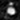} 
\includegraphics[width=0.0758\linewidth]{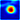} 
% ~
\includegraphics[width=0.0758\linewidth]{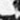} 
\includegraphics[width=0.0758\linewidth]{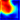} 
% ~
\includegraphics[width=0.0758\linewidth]{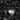} 
\includegraphics[width=0.0758\linewidth]{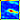} 
% ~
\includegraphics[width=0.0758\linewidth]{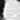} 
\includegraphics[width=0.0758\linewidth]{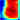} 
% ~
\includegraphics[width=0.0758\linewidth]{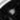} 
\includegraphics[width=0.0758\linewidth]{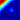} 
% ~
\includegraphics[width=0.0758\linewidth]{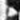} 
\includegraphics[width=0.0758\linewidth]{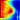} 
\\

% \noindent{~~~~~~1~~~~~~~~~~~~2~~~~~~~~~~~~3~~~~~~~~~~~~4~~~~~~~~~~~~5~~~~~~~~~~~~6~~~~~~~~~~~~7~~~~~~~~~~~~8~~~~~~~~~~~~9~~~~~~~~~~~10~~~~~~~~~~~11~~~~~~~~~~12 \hfill} \\

\end{tabular}
\caption{
Examples of random pulmonary nodules (odd columns) and their correspondent feature maps (even columns) from different levels of spatial context: long range context (top), medium range context (middle), and short range context (bottom).
Note that the feature maps are selected randomly from the second convolutional layer in the false positive reduction network.
We notice that exploiting information from different input scales (levels of context) assists the network in 
integrating spatial contextual information of different levels.
}
\label{fig:nodule_contextual_levels}
\end{figure*}

First, a suspected region (i.e. candidate locations proposed by the candidate proposal stage) is extracted in three different scales, such that each patch spans a different level of spatial context (area) around the center of the suspected region.
Here, we use small (15 x 15 x 15), medium (25 x 25 x 25), and large (40 x 40 x 40) patches, representing short, medium, and long range spatial context, respectively. Note that these are picked with respect to nodule size distribution.
% 
% 
% The extracted patches are then fed into three parallel convolutional
% \textcolor{violet}{
The extracted patches are then resized (individually) into 20 x 20 x 20 using bicubic interpolation, and are
% }
fed into three parallel convolutional
layers
% , 
in which each layer is specialised in a particular type of patches (i.e. small, medium, or large). The resulting feature maps are then individually down-sampled using a max pooling operation, aggregated (concatenated), and 
are jointly analysed by a sequence of four residual blocks \cite{method:resnet}, in which 2, 3, 3, and 3 residual units are incorporated, respectively. 
Moreover, the first residual block is followed by subsequent max pooling layer to further reduce the dimensionality of the 3D feature maps. 
In line with the concept of attention, our design aims at
assisting the network in learning contextual information by independently analysing nodules at different extents of spatial information (scales),
while
modeling correlations between the different contexts by jointly analysing the aggregated feature maps.
Fig. \ref{fig:nodule_contextual_levels} shows examples of nodule images of different contextual levels and a random selection of their feature maps.
We notice that by jointly exploiting
inputs of different scales,
the network was able to
integrate information from multiple levels of spatial context (MLSC).
In our experiment (Section \ref{section:experiments}), we demonstrate the benefit of the joint analysis and compare it against the performance when single scale inputs are used.

\begin{table*}
\centering
\caption{
Candidate proposal ablation study: FROC at different numbers of false positives per scan obtained by different methods under comparison on a randomly selected keep-out fold from LUNA16 dataset. Here, CA and SA stand for channel attention and spatial attention, respectively.
The highest scores are highlighted in bold.
}

\label{table:ablation_study_candidate_proposal}
\resizebox{1.0\textwidth}{!}{
\begin{tabular}{lcccccccc}

\hline
FROC & Mean & 0.125 & 0.25 & 0.5 & 1.0 & 2.0  & 4.0 & 8.0  \\
\hline

% RPN (baseline) 
RPN & 0.772 & 0.576 & 0.669 & 0.754 & 0.790 & 0.846 & 0.878 & 0.890 \\

% Transformer
RPN+Transformer \cite{wang2021transbts} & 0.746 & 0.537 & 0.619 & 0.706 & 0.764 & 0.826 & 0.872 & 0.897 \\

% SE
RPN+CA \cite{hu2018squeeze} & 0.813 & 0.638 & 0.697 & 0.792 & \textbf{0.856} & 0.877 & \textbf{0.911} & 0.919 \\

% CBAM-full
RPN+SA+CA \cite{woo2018cbam} & 0.743 & 0.576 & 0.631 & 0.687 & 0.745 & 0.816 & 0.862 & 0.886 \\

% CBAM-spatial-gate 
RPN+SA \cite{woo2018cbam} & 0.719 & 0.507 & 0.602 & 0.688 & 0.725 & 0.801 & 0.843 & 0.869 \\

% CBAM-channel-gate 
RPN+CA \cite{woo2018cbam} & 0.779 & 0.584 & 0.679 & 0.753 & 0.807 & 0.847 & 0.884 & 0.902 \\

% T5+T9
RPN+proposed CA+SA
& 0.782 & 0.544 & 0.689 & 0.765 & 0.810 & 0.864 & 0.886 & 0.912 \\

% T9
RPN+proposed SA
& 0.784 & 0.575 & 0.670 & 0.750 & 0.813 & 0.860 & 0.900 & 0.923 \\

% T5-ReLU 
RPN+proposed CA
& \textbf{0.826} & \textbf{0.657} & \textbf{0.761} & \textbf{0.808} & 0.834 & \textbf{0.887} & 0.909 & \textbf{0.929} \\

\hline
\end{tabular}
} % resize box 
\end{table*}

\begin{figure}[t]
\centering

\includegraphics[width=0.8\linewidth]{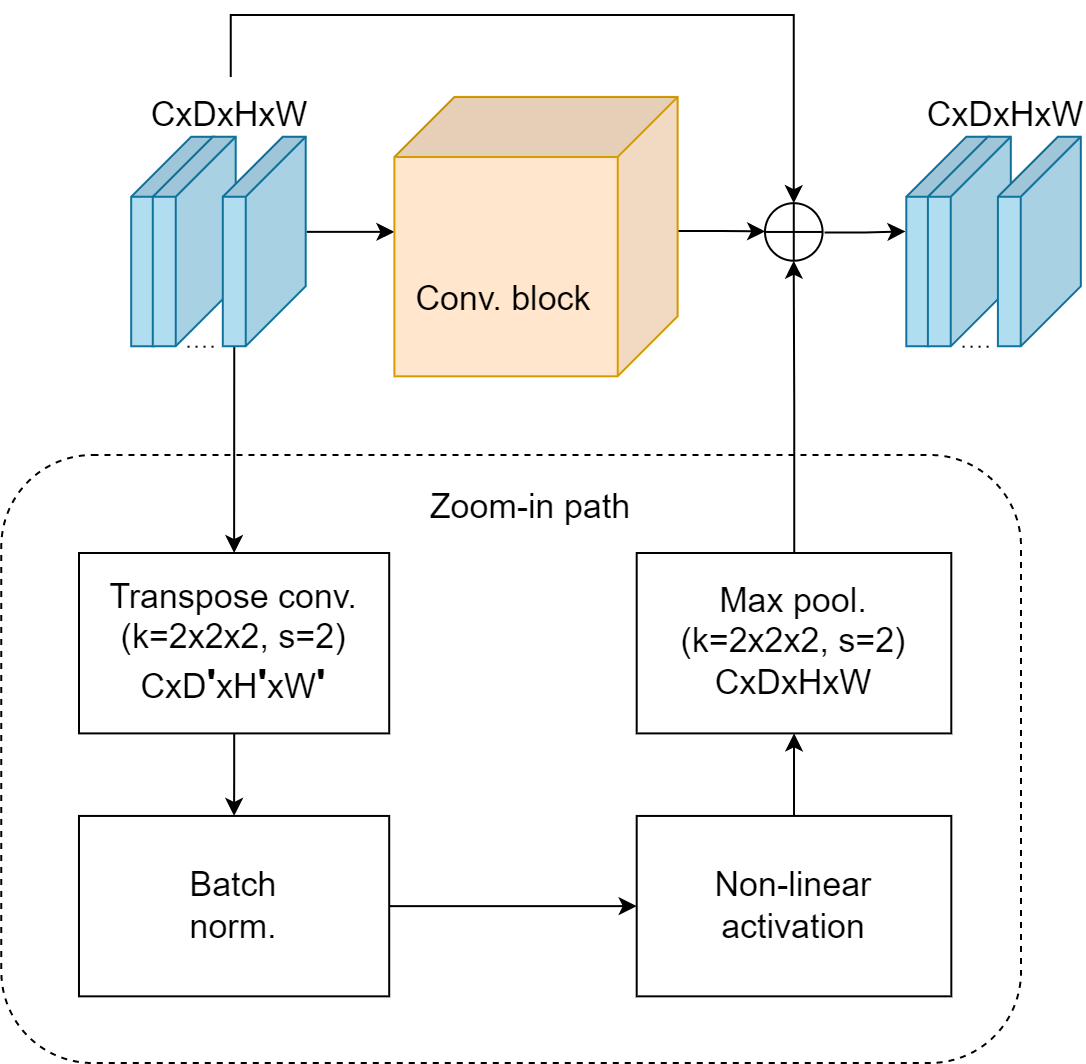}

\caption{
An overview of our proposed 3D fully convolutional zoom-in path within a residual block. Our zoom-in path takes intermediate feature maps of size $C\text{~x~}D\text{~x~}H\text{~x~}W$ as an input. These features are up-sampled by a transposed convolutional operation resulting a feature map of size $C\text{~x~}D'\text{~x~}H'\text{~x~}W'$ to assist the network in capturing finer details from different spatial scales. The extracted features are then normalised using a batch normalisation layer, mapped into a non-linear space, and are pooled using a max pooling operator in which the resulting embeddings are of the same dimensions as the original input.
These are then aggregated along with the features from the residual block using an element-wise addition process ($\bigoplus$ in the figure).
The parameters k and s represent the kernel size and the stride used in the convolutional and the max pooling layers.
}

\label{fig:zoominblock}
\end{figure}

As observed in Fig. \ref{fig:nodule_size_distribution_luna16}, most nodules are of small size, making the classification task more challenging
% \cite{larici2017lung,shen2021classification,das2006small,zhao2003automatic,wozniak2018small}
\cite{larici2017lung,shen2021classification,zhao2003automatic,wozniak2018small}. Moreover, smaller structures are prone to shrinkage due to the use of down-sampling layers, this contributes negatively to the quality of the acquired information, and therefore the accuracy of the detection.
Thus, we augment the residual units of our backbone network by zoom-in convolutional paths to assist the network in picking fine details from different feature scales.
An intermediate feature map F is first processed by a deconvolutional layer using kernels of size 2 x 2 x 2 and stride of 2, the feature map is therefore up-sampled by a factor of 2. The resulting embedding is then regularised using a batch normalisation layer, followed by 
a non-linear activation layer, and is finally 
transformed back into its original dimensions using a 3D max pooling operation.
The output of the zoom-in layer is incorporated as a skip connection similar to a residual path using element-wise addition.
See Fig. \ref{fig:zoominblock}. 
The intuition behind this strategy is to promote the network into learning to
emphasise (magnify) small structures using the up-sampling convolution, such that they are less prone to diminishing due to repeated down-sampling operations and the increasing receptive field. In our experiment (Section \ref{section:experiments}), we demonstrate the effectiveness of our proposed zoom-in path within the false positive reduction task, and evaluate its performance within the candidate proposal task.

Surprisingly, unlike our finding in the candidate proposal stage, we observe no significant improvement when integrating our proposed spatial attention (Eqs. \ref{attention_spatial_1} to \ref{attention_spatial_5}) within the false positive reduction task. A similar pattern was found when evaluating spatial attention from \cite{woo2018cbam}.
Nonetheless, when evaluating channel attention within the false positive reduction task,
we observe an enhanced performance using the channel attention approach proposed in \cite{woo2018cbam}, while no significant yield was found by integrating our proposed technique (Eqs. \ref{attention_channel_1} to \ref{attention_channel_2}).
This may be due to the different complexity of the two tasks.
In Section \ref{section:experiments}, we provide an extensive analysis in which we compare different types and combinations of attention techniques.
Accordingly, we adopt the cross-channel attention approach from \cite{woo2018cbam} into the building blocks of our false positive reduction network. The over all channel attention process is described in Eq. \ref{attention_channel_cbam}.
Similar to \cite{woo2018cbam}, channel attention is placed prior to the residual path in a residual unit.

The output of the last residual block is then passed into a fully connected notwork of 3 subsequent layers in which the final layer predicts nodule probabilities using a sigmoid function.
Similar to the candidate proposal stage, we evaluate focal cross entropy \cite{lin2017focal} within our false positive reduction task. 
We observe that setting the modulation parameter $\gamma$ (Eq. \ref{focal_loss}) to 0 produces the best overall performance. While using higher values
provoke lower false positive rates on the account of lower sensitivity in the nodule class, setting $\gamma$ to 0 induces a favorable balance in the overall performance.
Note that when $\gamma$ is set to 0, the loss computation is equivalent to using pure cross entropy.

\subsection{Integration of detection stages}
\label{section:Integrationofdetectionstages}

Our framework performs the detection in two stages, candidate proposal and false positive reduction. 
In line with \cite{dataset:LUNA16} and \cite{tang2018automated}, we find that ensembling the two stages such that the final prediction is a function of the two models, achieves the best detection performance.
First, for each tested image, predictions are then sorted in a descending order with respect to their nodule probability, locations with the highest 300 nodule scores are used as an initial pool of candidates. These are then evaluated against a threshold $t$, where $t$ is set to 0.3.
To eliminate any overlapping predictions, candidates are then processed by a non maximum suppression (NMS) operation with an intersection over union threshold of 0.1.

The remaining nodule candidates are then passed into the false positive reduction stage, where candidate regions are extracted and are processed by the false positive reduction network. For each detection, the final score is defined as the average probability of both stages. Last, detections with probabilities $\geq$ 0.3 are used to form the final set of detection.

\section{Experiments}
\label{section:experiments}

All experiments were implemented using PyTorch DL library with an NVIDIA V100 16GB GPU.
Both the candidate proposal and false positive reduction
stages were trained
% \textcolor{violet}{
(separately)
% } 
for 250 and 10 epochs ($\sim$2 and $\sim$0.5 days), respectively, using a batch size of 7 and 64, respectively.
For both stages, we use Stochastic Gradient Descent (SGD)
% \cite{paper:sgd} 
optimisation with an initial learning rate of 0.01. For the candidate proposal stage, the learning rate is decreased to 0.001, 0.0005, and 0.0001 after 50, 100, and 150 epochs, respectively, while it is conserved at 0.01 for the false positive reduction stage.
Network parameters are initialised using He et al. \cite{he2015delving}.

% We train and evaluate both stages following 10-fold cross validation as suggested by LUNA16 \cite{dataset:LUNA16}. 
% % 
% Performance is evaluated using the LUNA16's official metric, Free Response Receiver Operating Characteristic (FROC)
% % \cite{method:CPM}
% , in which sensitivity is computed at 7 predefined false positive rates (i.e. 0.125, 0.25, 0.5, 1, 2, 4, and 8) per scan. Most clinical setups define their effective threshold between 1 and 4 false positives per scan. Including lower false positive rates in the evaluation metric makes the task more challenging \cite{dataset:LUNA16}.

We train and evaluate both stages following 10-fold cross validation as suggested by LUNA16 \cite{dataset:LUNA16}. 
Performance is evaluated using Free Response Receiver Operating Characteristic (FROC) curve, 
in which sensitivity is computed at 7 predefined average false positive rates per scan (i.e. 0.125, 0.25, 0.5, 1, 2, 4, and 8), as well as Competition Performance Metric (CPM), the LUNA16's official metric, defined as the average sensitivity across all predefined false positive rates \cite{dataset:LUNA16}.
Most clinical setups define their effective threshold between 1 and 4 false positives per scan. Including lower false positive rates in the evaluation metric makes the task more challenging \cite{dataset:LUNA16}.

In the remainder of this section, we present data and pre-processing in Sections \ref{section:data} and \ref{section:preprocessing}, respectively. Furthermore, to evaluate the contribution of 
the individual components of
our approach,
we first perform an ablation study for each of the detection stages (candidate proposal and false positive reduction stage) in Section \ref{section:ablationstudy}, then we evaluate the performance of the fully integrated system in Section \ref{integratedsystemresults}.

% \begin{table}
% \centering
% \caption{
% FROC at different numbers of false positives per scan obtained by the best performing candidate proposal network (proposed RPN with cross-channel attention)
% on a randomly selected keep-out fold from LUNA16 dataset using different activation functions.
% }

% \label{table:activation_function_comparison}
% % \resizebox{!}{0.02492\paperheight}
% \resizebox{!}{0.034\paperheight}
% {
% \setlength{\tabcolsep}{0.2em} % horizontal padding
% {\renewcommand{\arraystretch}{1.2}%  vertical padding

% \begin{tabular}{lcccccccc}

% \hline
% FROC & Mean & 0.125 & 0.25 & 0.5 & 1.0 & 2.0  & 4.0 & 8.0  \\
% \hline
% % Candidate proposal stage:

% % T5 - ReLU: 
% % RPN+proposed CA
% ReLU & 0.826 & \textbf{0.657} & \textbf{0.761} & 0.808 & 0.834 & 0.887 & 0.909 & 0.929 \\

% % T5 - Leaky ReLU:
% % \textcolor{red}{RPN+proposed CA+Leaky ReLU}
% Leaky ReLU & 0.822 & 0.602 & 0.732 & \textbf{0.826} & 0.868 & 0.897 & 0.910 & 0.916 \\

% % T5 - modified ReLU:
% % RPN+proposed CA+modified ReLU
% prop. mod. ReLU & \textbf{0.833} & 0.642 & 0.733 & 0.808 & \textbf{0.882} & \textbf{0.909} & \textbf{0.922} & \textbf{0.938} \\

% \hline
% \end{tabular}
% } % horizontal and vertical padding
% } % resize box 
% \end{table}

\begin{table}
\centering
\caption{FROC at different numbers of false positives per scan obtained by the best performing candidate proposal network (proposed RPN with cross-channel attention) on a randomly selected keep-out fold from LUNA16 dataset using different activation functions.
}

\label{table:activation_function_comparison}
\resizebox{!}{0.032\paperheight}
{
\setlength{\tabcolsep}{0.2em} % horizontal padding
{\renewcommand{\arraystretch}{1.2}%  vertical padding

\begin{tabular}{lcccccccc}

\hline
FROC & Mean & 0.125 & 0.25 & 0.5 & 1.0 & 2.0  & 4.0 & 8.0  \\
\hline

% T5 - ReLU: 
% RPN+proposed CA
ReLU & 0.826 & \textbf{0.657} & \textbf{0.761} & 0.808 & 0.834 & 0.887 & 0.909 & 0.929 \\

% T5 - Leaky ReLU:
Leaky ReLU & 0.822 & 0.602 & 0.732 & \textbf{0.826} & 0.868 & 0.897 & 0.910 & 0.916 \\

% T5 - ELU:
ELU & 0.830  & 0.641 & 0.743 & 0.802 & 0.861 & 0.895 & \textbf{0.923} & \textbf{0.943} \\

% T5 - modified ReLU:
% RPN+proposed CA+modified ReLU
prop. mod. ReLU & \textbf{0.833} & 0.642 & 0.733 & 0.808 & \textbf{0.882} & \textbf{0.909} & 0.922 & 0.938 \\

\hline
\end{tabular}
} 
} 
\end{table}

% zoom-in block :

\subsection{Data}
\label{section:data}

We use the LUNA16\footnote{available at https://luna16.grand-challenge.org/} (LUng Nodule Analysis 2016) \cite{dataset:LUNA16} dataset, a subset of LIDC/IDRI dataset \cite{dataset:LIDC}, to carry out our experiments.
The LIDC/IDRI dataset was collected in two stages, a blinded annotation stage, where 4 radiologists were asked to independently mark suspicious locations, and an unblinded stage, where results of all radiologists where anonymised and provided to each of the radiologists to assist them and re-evaluate their initial annotations.

% As recommended by \cite{kazerooni2014acr}, \cite{naidich2013recommendations}, and \cite{manos2014lung}, thin slices must be used for pulmonary nodule analysis, therefore, 
LUNA16 excludes scans with slice thickness $>2.5$ mm. Scans with impaired slices or inconsistent slice spacing are also excluded, resulting a total of 888 CT scans.
Moreover, only nodules that are $\geq3$ mm and are accepted by a minimum of 3 out 4 radiologists are included in LUNA16, summing up to 1186 nodule labels. Following the lung cancer screening protocols
% in \cite{aberle16reduced}
, any remaining annotations are flagged as irrelevant findings.

Furthermore, for the false positive reduction task, LUNA16 provides 754,975 candidate locations acquired using multiple existing methods.
% (\cite{method:fpcandluna161, method:fpcandluna162, method:fpcandluna163, method:fpcandluna164, method:fpcandluna165}).
These include 1,166 nodules that match the radiologists annotations and 753,809 of non-nodule locations.

\subsection{Pre-processing}
\label{section:preprocessing}

Similar to \cite{method:deepseed} and \cite{method:deeplung}, we pre-process LUNA16 images by clipping their intensity values between -1200 and 600 Hounsfield units (HU) followed by rescaling them between 0 and 1. We use the lung masks provided by LUNA16 to isolate the informative lung region and mask out any surrounding organs. 
Due to GPU limitations, for the candidate proposal stage, images were split into patches of size 128 x 128 x 128 during the training stage. We follow the same approach for the testing stage allowing an overlap of 32 pixels between the cropped patches.

Additionally, we follow a cross-sectional augmentation strategy, where scans, along with their annotations, are transformed from axial plane (original plane), into sagittal and coronal planes.

For the false positive reduction stage,  
patches are extracted using annotation coordinates provided by LUNA16 for the false positive reduction task. Each patch is cropped in 3 different sizes (levels of spatial context), 15 x 15 x 15, 25 x 25 x 25, and 40 x 40 x 40. This ensures the coverage of 99\% of the nodules \cite{method:dou2016multilevel}. All patches are then resized to 20 x 20 x 20 using bicubic interpolation.
Furthermore,
% \textcolor{violet}{
during training,
% }
we augment the nodule class using cross-sectional augmentations, followed by 1-pixel shifts along the z, y, and x axes.

\begin{table*}
\centering
\caption{
False positive reduction ablation study: FROC at different numbers of false positives per scan obtained by different methods under comparison on a randomly selected keep-out fold from LUNA16 dataset.
The top section of the table compares results of the false positive reduction network when incorporating spatial information from different contextual levels. The bottom section presents a comparison of different attention mechanisms.
Here, CA and SA stand for channel attention and spatial attention, respectively.
The highest scores are highlighted in bold.}

\label{table:ablation_study_fp_reduction}
\resizebox{1.0\textwidth}{!}{
\begin{tabular}{lcccccccc}

\hline
FROC & Mean & 0.125 & 0.25 & 0.5 & 1.0 & 2.0  & 4.0 & 8.0  \\
\hline

Short range spatial context & 0.623 & 0.207 & 0.354 & 0.538 & 0.681 & 0.763 & 0.877 & 0.936 \\

Medium range spatial context & 0.769 & 0.503 & 0.646 & 0.738 & 0.822 & 0.861 & 0.902 & 0.910 \\

Long range spatial context & 0.715 & 0.487 & 0.578 & 0.681 & 0.740 & 0.814 & 0.854 & 0.855 \\

Multi-level spatial context (MLSC) & 0.792 & 0.529 & 0.626 & 0.738 & \textbf{0.831} & \textbf{0.915} & \textbf{0.946} & \textbf{0.963} \\

MLSC+zoom-in (MLSC-Z) & \textbf{0.813} & \textbf{0.661} & \textbf{0.730} & \textbf{0.783} & 0.822 & 0.846 & 0.912 & 0.938 \\

\hline

MLSC-Z+Transformer \cite{wang2021transbts}
& 0.672 & 0.260 & 0.453 & 0.631 & 0.714 & 0.840 & 0.902 & 0.904 \\

MLSC-Z+proposed CA+SA 
& 0.678 & 0.455 & 0.500 & 0.575 & 0.715 & 0.792 & 0.844 & 0.866 \\

MLSC-Z+proposed SA & 0.805 & 0.564 & 0.661 & 0.790 & 0.828 & 0.881 & 0.939 & \textbf{0.972} \\

MLSC-Z+proposed CA
& 0.775 & 0.564 & 0.640 & 0.740 & 0.790 & 0.849 & 0.899 & 0.946 \\

MLSC-Z+CA+SA \cite{woo2018cbam} & 0.740 & 0.433 & 0.566 & 0.713 & 0.801 & 0.849 & 0.907 & 0.909 \\

MLSC-Z+SA \cite{woo2018cbam}
& 0.778 & 0.505 & 0.634 & 0.744 & 0.821 & 0.903 & 0.919 & 0.919 \\

MLSC-Z+CA \cite{hu2018squeeze} & 0.833 & 0.593 & 0.711 & 0.812 & \textbf{0.883} & \textbf{0.926} & \textbf{0.950} & 0.957 \\

MLSC-Z+CA \cite{woo2018cbam} & \textbf{0.848} & \textbf{0.702} & \textbf{0.745} & \textbf{0.815} & 0.862 & 0.906 & 0.940 & 0.963 \\

\hline
\end{tabular}
} % resize box 
\end{table*}

\subsection{Ablation study}
\label{section:ablationstudy}

\subsubsection{Candidate proposal stage}
\label{section:candidateproposalresults}

\begin{figure}
\centering

\includegraphics[width=0.99\linewidth]{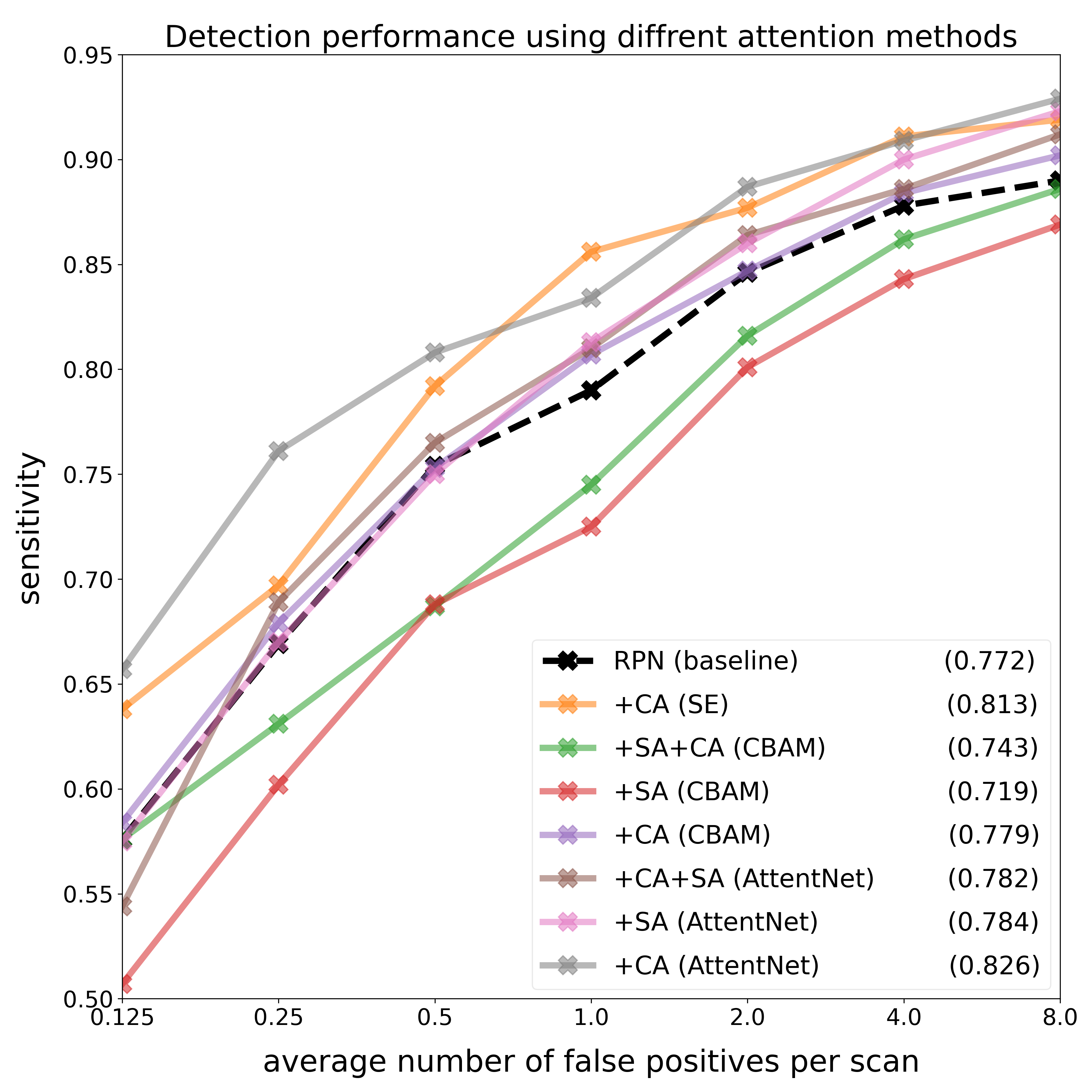}

\caption{FROC of the proposed region proposal network using different attention mechanisms under comparison on a randomly selected keep-out fold from LUNA16 dataset. Legend indicates CPM score for each tested detector.}

\label{fig:attention_froc_comparison}
\end{figure}

For the candidate proposal task, we compare the impact of different attention mechanisms against the performance of our backbone CNN when no attention is used. Results are presented in Table \ref{table:ablation_study_candidate_proposal}, and Fig. \ref{fig:attention_froc_comparison}.
Performing 10-folds cross validation requires extensive amounts of time
(e.g., $\sim$2 days per fold), therefore, 
for the purpose of evaluating the candidate proposal stage, 
training and testing were performed using a randomly selected keep-out fold.
Nonetheless, in Section \ref{integratedsystemresults}, we evaluate our fully integrated system using 10-folds cross validation as suggested in LUNA16 \cite{dataset:LUNA16}.
Note that all our experiments are trained using ReLU activation function unless stated otherwise.

In our first experiment, we evaluate the performance of our candidate proposal network when no attention is incorporated. This will serve as a baseline for our candidate proposal stage. We find that the network has successfully detected lung nodules at a recall (sensitivity) of 0.878 at 4 false positives per scan, and 
% an FROC score
a CPM score (i.e. average sensitivity over 7 false positive thresholds, 0.125, 0.25, 0.5, 1.0, 2.0, 4.0, 8.0) of 0.772, prior to any false positive reduction.
Note that most clinical setups define their effective threshold between 1 and 4 false positives per scan \cite{dataset:LUNA16}.
This demonstrates a compelling potential of CNNs in the pulmonary nodule detection task.

% \begin{figure}[t]
% \centering

% \includegraphics[width=0.85\linewidth]{figures/activation_functions/valdiation_loss_ReLU_vs_LeakyReLU_TanHReLU_1.1.png}

% \caption{
% Validation loss of the proposed lung nodule detection network using different activation functions.
% Our proposed modified ReLU activation is smooth around the origin, promoting faster learning in contrast to both, pure ReLU and leaky ReLU activations.}

% \label{fig:val_loss_comparison_of_activations}
% \end{figure}

\begin{figure}[t]
\centering

\includegraphics[width=0.95\linewidth]{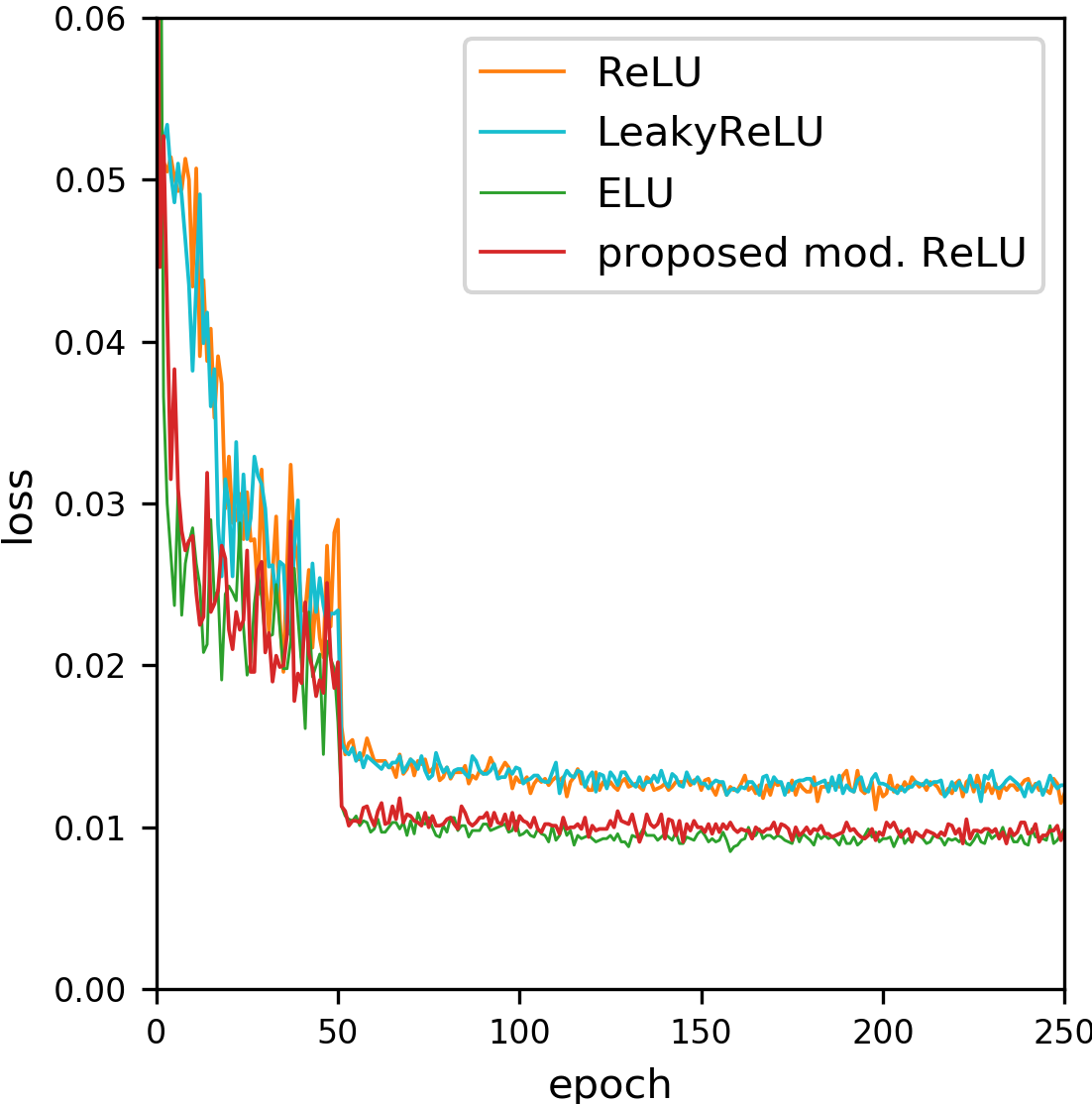}

\caption{Validation loss of the proposed lung nodule detection network using different activation functions.
Our proposed modified ReLU activation, as well as ELU activatons, are smooth around the origin, promoting faster learning in contrast to, pure ReLU and leaky ReLU activations.}

\label{fig:val_loss_comparison_of_activations}
\end{figure}

To evaluate the impact of different attention mechanisms, we compare
cross-channel and spatial-wise attention when applied individually, and when applied in combination (i.e. subsequently) within the building blocks of the detection network.

We find that all channel attention approaches (i.e. the proposed approach and the ones from \cite{woo2018cbam} and \cite{hu2018squeeze}) contribute positively to the overall detection performance, with our proposed strategy producing the highest 
% FROC score 
CPM score
of 0.826, in contrast to 0.779 and 0.813 using the attention approaches from \cite{woo2018cbam} and \cite{hu2018squeeze}, respectively.
This demonstrates the advantage of our fully convolutional design;
by replacing the MLP units (i.e. channel attention in \cite{woo2018cbam} and \cite{hu2018squeeze}) with convolutional operations, 
we avoid the need for heavy dimesionality reduction (see Section \ref{section:candidateproposalstage}),
and attention can therefore be efficiently performed using spatial embeddings of higher dimensions, leading to an enhanced performance.
As demonstrated by the results, this is particularly useful when managing 3D data of high spatial resolution.

Moreover, when evaluating the impact of spatial attention, we find that our proposed spatial attention strategy
was able to successfully enhance the network's performance with 
% an FROC score
a CPM score
of 0.784, showing an increase of 1.2\% comparing to the baseline network when no attention is used.
On the other hand, we find that spatial approach from \cite{woo2018cbam} leads to a worse performance comparing to the baseline network.
This shows that,
in contrast to \cite{woo2018cbam}, where spatial descriptors are aggregated using element-wise pooling operations along the channel axis, neglecting the volumetric aspect of the data, the incorporation of cross-sectional spatial information in our attention strategy can successfully assist the network in learning correlations between the different image cross-sections, and consequently capturing important 3D information within the images.

Furthermore, we experiment with multi-headed attention (Transformer) \cite{vaswani2017attention,dosovitskiy2020image} to evaluate the impact of incorporating long-range correlations and global context on the pulmonary nodule detection task.
For this experiment, we adopt a hybrid architecture that takes advantage of the spatial representational power of CNNs and the ability in modeling long-range dependencies in Transformers \cite{wang2021transbts}.
Particularly, we use our 3D encoder CNN to extract and down-sample pulmonary nodule 
feature maps
in which a Transformer takes as an input to analyse and perform self-attention
Subsequently, the resulting feature maps are passed into the decoder part of the CNN where the final detection is performed.
To the best of our knowledge, we are the first to evaluate Transformer networks within the pulmonary nodule detection task.
We find that Transformer based detector produces an increased sensitivity of 0.897 at 8 false positives per scan comparing to 0.890 using the baseline network when no attention is used. However, Transformer shows a decrease of 2.6\% in the overall 
% FROC score 
CPM score
in contrast to the baseline network.
Unlike CNNs, the tokenisation process of input images in Transformer leads to a limited ability in modeling
local structures,
and consequently, a limited translation equivariance \cite{dosovitskiy2020image}.
This may indicate that the pulmonary data benefits more from localised
attention (e.g. convolution based attention) than long range global attention.
In \cite{dosovitskiy2020image}, the authors suggest that training Transformers using larger datasets (e.g. 14M-300M images) may help in overcoming this issue, however,
due to the lack of a suitable dataset (e.g. a dataset of similar configurations and nature to the lung images) of this size,
this remains a challenge when dealing with pulmonary images.

\begin{table*}
\centering
\caption{
FROC at different numbers of false positives per scan obtained by our fully integrated two stage pulmonary nodule detection system, AttentNet, in contrast to baseline methods using 10-folds cross validation on LUNA16 dataset.
Here, TTA indicates the use of testing time augmentation. The highest scores are highlighted in bold.
}

\label{table:integrated_system_vs_baselines}
\resizebox{1.0\textwidth}{!}{
\begin{tabular}{lcccccccc}

\hline
FROC & Mean & 0.125 & 0.25 & 0.5 & 1.0 & 2.0  & 4.0 & 8.0 \\
\hline

% V-Net
ZNET \cite{berens2016znet} & 0.811 & 0.661 & 0.724 & 0.779 & 0.831 & 0.872 & 0.892 & 0.915 \\

% 3D Faster RCNN 
3D RCNN \cite{liao2019evaluate} & 0.834 & 0.662 & 0.746 & 0.815 & 0.864 & 0.902 & 0.918 & 0.932 \\

% DeepLung
DeepLung \cite{method:deeplung} & 0.842 & 0.692 & 0.769 & 0.824 & 0.865 & 0.893 & 0.917 & 0.933 \\

% DeepSeed 
DeepSeed 
\cite{method:deepseed} & 0.862 & 0.739 & 0.803 & 0.858 & 0.888 & 0.907 & 0.916 & 0.920 \\

% Ours (No FP reduction) NTTA
AttentNet (RPN) (Ours) & 0.842 & 0.656 & 0.774 & 0.831 & 0.874 & 0.903 & 0.923 & 0.936 \\

% Ours NTTA
AttentNet (Ours) & 0.871 & \textbf{0.752} & \textbf{0.817} & \textbf{0.857} & 0.885 & \textbf{0.920} & 0.933 & 0.933 \\

% Ours TTA
AttentNet+TTA (Ours)  & \textbf{0.874} & 0.748 & 0.812 & 0.856 & \textbf{0.893} & 0.919 & \textbf{0.942} & \textbf{0.945} \\

\hline

% Accurate Pulmonary Nodule Detection in Computed Tomography Images Using Deep Convolutional Neural Networks (APNDCTDCNN)
% we propose a novel pulmonary nodule detection CAD system based on deep convolution networks, where a deconvolutional improved
% Faster R-CNN is developed to detect nodule candidates from axial slices
% and a 3D DCNN is then exploited for false positive reduction. Experimental results on the LUNA16 Nodule Detection Challenge demonstrate that
% the proposed CAD system ranks the 1st place of Nodule Detection Track
% (NDET) with an average FROC-score of 0.891.
APNDCTDCNN \cite{method:accurateluna}  & 0.891 & - & - & - & - & -& - & - \\

% 3D Region Proposal U-Net with Dense and Residual Learning for Lung Nodule Detection
% Since the image volume is too large to fit into a standard GPU for CNN processing directly even after lung
% segmentation. A window-based detection approach is employed, where a 3D window slides over the lung
% region and a cubic patch is extracted each time, fed to the 3D RPN for nodule detection. Figure 1 illustrates
% the detection framework, which is a dual-path network consisting of two U-shaped sub-networks: one follows
% the DenseNet paradigm while the other follows the ResNet paradigm. A label fusion is applied using nonmaximal suppression (NMS) method [12]. The neural network implementation was based on the popular
% PyTorch framework [18]. Further details may not be published at this point due to paper preparation, and will
% be released later.
% The experiments were performed on a standard machine with two NVIDIA K80 GPUs, where network
% training took around a week for each fold of cross validation. At test time, end-to-end workflow took less
% than 1 minute, among which, lung segmentation took around 20~30s and nodule detection by network
% inference took around 10~20s, depending on the size of CT image
3DDUALRPN \cite{Xie2017}  & \textbf{0.922} & \textbf{0.801} & \textbf{0.860} & \textbf{0.919} & \textbf{0.952} & \textbf{0.968} &\textbf{ 0.977} & \textbf{0.977}\\

\hline
\end{tabular}
} % resize box 
\end{table*}

Generally,
while both our spatial and cross-channel attention demonstrate a positive impact within the pulmonary nodule detection task, we observe that the network yields a better performance using channel-wise attention in contrast spatial attention. A similar trend is observed when evaluating channel and spatial attention from \cite{woo2018cbam} and \cite{hu2018squeeze}.
Moreover, when combining both our channel and spatial attention by applying them in subsequently (as suggested in \cite{woo2018cbam}), we notice an increase in the 
% FROC score 
CPM score
comparing to the baseline network when no attention is used, however,
we find that the network benefits the most when
cross-channel and spatial attention are incorporated 
individually, with channel attention being the best performing amongst all configurations.
In the same line, when evaluating the attention methods from \cite{woo2018cbam}, we find that their channel attention produces better results in contrast to both, their spatial attention, and spatial and channel attention when applied in combination.
These observations indicate a high importance of inter-channel dependencies in the pulmonary nodule detection task.
In fact, we argue that channel attention not only assists the network in modelling inter-channel correlations,
but also inherently infers spatial attention by assisting the network
in focusing on important feature maps (channels) in which informative spatial features are embedded.
Accordingly, and due to the improved performance demonstrated in our experiment, we continue using our proposed channel attention as the method of choice for the candidate proposal task.

In our preliminary experiments, we evaluate our zoom-in path within the candidate proposal stage. We observe comparable performance to the network when no zoom-in paths are used.
This is expected since the candidate proposal network benefits from an encoder-decoder design in which spatial features are extracted from multiple spatial scales.
However, the zoom-in path demonstrates benefits when incorporated within the false positive reduction task, where an encoder-decoder design is not straightforwardly applicable due to the relatively narrow input dimensions (details in Section \ref{section:falsepositivereductionresults}).

% \sout{Finally, we evaluate the impact of our proposed modified ReLU activation function within the nodule detection task by applying it to the
% best performing
% network configuration amongst all tested detectors (i.e. RPN with the proposed cross-channel attention), and compare its performance to the network with pure ReLU and Leaky ReLU activations.
% Results are presented in Table \ref{table:activation_function_comparison}.
% % 
% We find that the modified ReLU activation produces an increased 
% % FROC score
% CPM score
% of 0.833 comparing to 0.826 and 0.822 when using pure ReLU or Leaky ReLU. Specifically, a significant increase is observed in the
% sensitivity at false positive per scan thresholds from 1.0 to 8.0, in which an effective clinical threshold (i.e. 1.0 - 4.0) is defined within \cite{dataset:LUNA16}.
% % 
% This demonstrates that allowing small outputs for inputs in the negative range of ReLU can improve the gradient flow and therefore enhances the learning process, see Figs. \ref{fig:activiation_functions} and \ref{fig:val_loss_comparison_of_activations}.
% % 
% Moreover, unlike Leaky ReLU, our modified ReLU activation is bounded for negative inputs, promoting network regularisation and reducing the risk of overfitting.
% Therefore, we adopt this 
% modified ReLU activation
% for our detection network and continue using it for the remainder of our experiment.}

Finally, we evaluate the impact of our proposed modified ReLU activation function within the nodule detection task by applying it to the
best performing
network configuration amongst all tested detectors (i.e., RPN with the proposed cross-channel attention), and compare its performance to the network with pure ReLU, Leaky ReLU, and ELU activations.
Results are presented in Table \ref{table:activation_function_comparison}.
We find that the modified ReLU activation produces an increased FROC score of 0.833 comparing to 0.826, 0.822, and 0.830 when using pure ReLU, Leaky ReLU, or ELU activations, respectively.
More particularly, when comparing to pure ReLU and Leaky ReLU, a significant increase is observed in the
sensitivity at false positive per scan thresholds from 1.0 to 8.0, in which an effective clinical threshold (i.e., 1.0 - 4.0) is defined within \cite{dataset:LUNA16}.
In the same line, the proposed modified ReLU activation achieves the highest sensitivity at 1 and 2 false positives per scan. ELU on the other hand obtains higher sensitivity at 4 and 8 false positives per scan, and a comparable average FROC score to the proposed activation.
This is expected since ELU and the proposed activation have similar characteristics
over the negative segment of the activation function.
The performance gain using the proposed activation may be explained by the smoother gradient flow in contrast to that in ELU as demonstrated in 
% Fig. \ref{fig:activiation_functions}.
% see 
Figs. \ref{fig:activiation_functions} and \ref{fig:val_loss_comparison_of_activations}.
Unlike unbounded activations (e.g., Leaky ReLU), our modified ReLU activation, as well as ELU activation, are bounded for negative inputs, promoting network regularisation and reducing the risk of overfitting.
These observations demonstrate that allowing small outputs for inputs in the negative range of ReLU can improve the gradient flow and therefore enhances the learning process, see Fig. \ref{fig:val_loss_comparison_of_activations}.
Accordingly, we adopt this 
modified ReLU activation
for our detection network and continue using it for the remainder of our experiment.

\begin{figure}[!t] 
\centering
\begin{tabular}{ccc}

\includegraphics[width=0.305\linewidth]{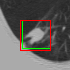} 
\includegraphics[width=0.305\linewidth]{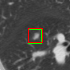} 
\includegraphics[width=0.305\linewidth]{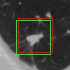} \\
% \text{~Slice 165 (165)~~~~~~~Slice 98 (99)~~~~~~Slice 252 (252)} \\

\includegraphics[width=0.305\linewidth]{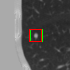} 
\includegraphics[width=0.305\linewidth]{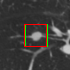} 
\includegraphics[width=0.305\linewidth]{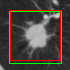} \\
% \text{~Slice 110 (110)~~~~~~~Slice 141 (141)~~~~~~Slice 183 (181)} \\

\includegraphics[width=0.305\linewidth]{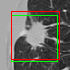} 
\includegraphics[width=0.305\linewidth]{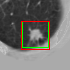} 
\includegraphics[width=0.305\linewidth]{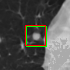} \\
% \text{~Slice 153 (150)~~~~~~~Slice 145 (146)~~~~~~Slice 128 (128)} \\

\includegraphics[width=0.305\linewidth]{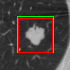} 
\includegraphics[width=0.305\linewidth]{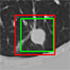} 
\includegraphics[width=0.305\linewidth]{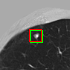} \\
% \text{~Slice 235 (233)~~~~~~~Slice 64 (54)~~~~~~Slice 182 (181)} \\

\includegraphics[width=0.305\linewidth]{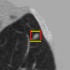} 
\includegraphics[width=0.305\linewidth]{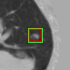} 
\includegraphics[width=0.305\linewidth]{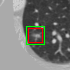} \\
% \text{~Slice 99 (99)~~~~~~~Slice 148 (148)~~~~~~Slice 54 (55)} \\

\includegraphics[width=0.305\linewidth]{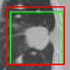} 
\includegraphics[width=0.305\linewidth]{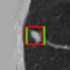} 
\includegraphics[width=0.305\linewidth]{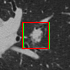} \\
% \text{~Slice 71 (69)~~~~~~~Slice 108 (107)~~~~~~Slice 175 (175)} \\

\end{tabular}

\caption{
Pulmonary nodules detected by AttentNet (green) and their correspondent ground-truth boxes (red).
% Note that for visualisation purpose, we show the center slices of the detected nodules.
% Numbers below each image indicate detected nodule center, and (in brackets) ground-truth centers.
}
\label{fig:nodule_detection_samples}
\end{figure}

\subsubsection{False positive reduction stage}
\label{section:falsepositivereductionresults}

For the false positive reduction task, we first evaluate the impact of using different input scales (levels of spatial context) on the detection, and compare it against our proposed joint analysis approach in which we incorporate inputs of multi-level spatial context (MLSC). We then evaluate the performance of our zoom-in path, as well as different attention mechanisms, against the performance of our backbone CNN when no attention is used.
Results are presented in Table \ref{table:ablation_study_fp_reduction}.
Similar to the candidate proposal stage, we evaluate our false positive reduction network using a randomly selected keep-out fold.

When comparing different input scales for the false positive reduction network, we find that medium range spatial context (25 x 25 x 25) produces the best performance in contrast to short (15 x 15 x 15) and long (40 x 40 x 40) range spatial context. However, we find that jointly analysing MLSC inputs enhances the overall detection comparing to using any of input scales individually,
with 
% an FROC score
a CPM score
of 0.792
comparing 0.769 produced using medium range spatial context inputs.
This demonstrates the importance incorporating spatial information from different levels when managing
objects with the high size variability such as the pulmonary nodules (see Fig. \ref{fig:nodule_size_distribution_luna16}).
Furthermore, Fig. \ref{fig:nodule_contextual_levels} shows that our MLSC network was able to
integrate spatial contextual information of different levels using our joint analysis approach.
We also find that augmenting our multi-level spatial context network with the proposed zoom-in path (MLSC-Z)
enhances the performance even further,
with 
% an FROC score
a CPM score
of 0.813, showing an increase of 2.1\% comparing to the performance when the zoom-in path is not used.
% 
% Note that when evaluating our zoom-in path within the candidate proposal stage, we observe comparable performance to the network when no zoom-in is used. This is expected since our candidate proposal network benefits from an encoder-decoder design in which feature maps are extracted from multiple spatial scales.
% 
% On the other hand, 
% 
Note that unlike the candidate proposal stage where the network benefits from an encoder-decoder design in which feature maps are extracted from multiple spatial scales, 
due to the nature of the false positive reduction task and the relatively narrow input dimensions, an encoder-decoder approach is not straightforwardly applicable. Thus, we design and incorporate our zoom-in path within the building blocks of the false positive reduction network.

When comparing different attention approaches, similar to our observation in the candidate proposal stage 
(Section \ref{section:candidateproposalresults})
, channel attention demonstrates the best performance comparing to both, spatial and global (i.e. self-attention) attention.
Indeed, the essence of cross-channel attention is in line with the concept of our joint analysis approach, in which we aim to assist the network in capturing correlations between inputs of different levels of spatial context, this suggests that both, channel attention and the joint analysis approach provide complementary information to one another.
Furthermore, contrary to our finding in the candidate proposal task, we notice that channel attention approaches form \cite{woo2018cbam} and \cite{hu2018squeeze} show better performance comparing to our proposed channel attention within the false positive reduction task, with the channel attention from \cite{woo2018cbam} producing the highest results, showing an increase of 3.5\% in the overall 
% FROC score
CPM score
in contrast to the performance when no attention is used.
This may be due the different complexity between the two tasks, this may indicate that the MLP based attention (e.g. \cite{woo2018cbam} and \cite{hu2018squeeze}) is more suitable for the false positive reduction task in contrast to our fully convolutional channel-wise attention strategy.
Accordingly, we continue using channel attention from \cite{woo2018cbam} as our approach of choice in the false positive reduction stage.
On the other hand, we find that our spatial attention produces the highest sensitivity
at 8 false positives per scan,
and a higher overall performance in contrast to the spatial attention approach from \cite{woo2018cbam}. This demonstrates the benefits of incorporating cross-sectional spatial information in our spatial attention comparing to the channel-wise pooling approach used in \cite{woo2018cbam}.

Overall, our results demonstrate the importance of exploiting morphological information when dealing with pulmonary nodules. Combining nodule morphology with attention mechanisms further enhances the network ability in learning effective embeddings and consequently produce more robust predictions.

\subsection{Integrated system performance}
\label{integratedsystemresults}

In this section, we evaluate the performance of the proposed nodule detection system by integrating both detection stages (candidate proposal and false positive reduction stage) in an ensemble model. To allow comparison with existing methods, all experiments in this section are done following 10-folds cross validation as suggested in LUNA16 \cite{dataset:LUNA16}. Results are presented in Table \ref{table:integrated_system_vs_baselines}.

First, we evaluate the performance of our candidate proposal network (RPN) when no false positive reduction stage is incorporated.
We find that that our network produces a better overall performance in contrast to the two stage detector \cite{berens2016znet} and higher or comparable results to the single stage detectors, \cite{liao2019evaluate}, \cite{method:deeplung} and \cite{method:deepseed}.
Additionally, when comparing our network to baselines at false positives per scan $\geq 1$, in which an effective clinical threshold is defined \cite{dataset:LUNA16}, we find that our network outperforms \cite{berens2016znet}, \cite{liao2019evaluate}, and \cite{method:deeplung} at 1.0, 2.0, 4.0 and 8.0 false positives per scan, as well as \cite{method:deepseed} at 4.0 and 8.0 false positives per scan with a comparable performance at 1.0 and 2.0 false positives per scan.
Note that our proposed network contains 3.1M trainable parameters, in contrast to 17M, 5.4M, 1.4M, and 5.4M parameters in \cite{berens2016znet}, \cite{liao2019evaluate}, \cite{method:deeplung}, and \cite{method:deepseed}, respectively, making it one of the most compact networks amongst the methods under comparison.
This demonstrates the positive impact of our proposed 3D attention strategy, by enabling the network in focusing on important features, our network was able efficiently to produce robust detections, prior to any false positive reduction.

Furthermore, we find that by integrating both detection stages as an ensemble, our network was able to achieve
% the 
% highest 
higher performance score comparing to 3D detector methods \cite{liao2019evaluate}, \cite{method:deeplung}, and \cite{method:deepseed}, 
% amongst all methods under comparison.
with 
% an FROC score
a CPM score
of 0.871, showing a significant increase of 2.9\% in contrast to our network prior to the false positive reduction step, and 0.9\% comparing to \cite{method:deepseed}, in which MLP based SE attention blocks were utilised.
% , the second highest nodule detector.
% 
This indicates the importance of the false positive reduction step, particularly when managing a critical task such as pulmonary nodule detection. By analysing suspected nodule regions and extracting deeper semantic features, the network was able was able to produce more robust predictions.

On the other hand, both APNDCTDCNN \cite{method:accurateluna} and 3DDUALRPN \cite{Xie2017} demonstrate higher performance on the LUNA16 dataset compared to our attention-based approach, achieving CPM scores of 0.891 and 0.922, respectively. However, it is important to note that \cite{method:accurateluna} relies on Faster R-CNN \cite{ren2015faster} for their candidate proposal task, which involves 41M trainable parameters, in contrast to our candidate proposal stage, which consists of only 3.1M parameters. Meanwhile, 3DDUALRPN \cite{Xie2017} uses a dual-path network, consisting of a ResNet50 based RPN as well as a DenseNet path, in an ensemble manner for their candidate detection stage, resulting in a optemising time of 1 week on an NVIDIA K80 12GB GPU (4992 CUDA cores) for each fold of cross-validation. The exact number of trainable parameters for \cite{Xie2017} was not reported. In contrast, our model was trained using all 10 cross-validation folds in just 2 days on an NVIDIA V100 16GB GPU (5120 CUDA cores), making it significantly more efficient in terms of training time and resource usage. While this is an indirect comparison, it effectively highlights the efficiency of our approach.
% Accordingly, we suggest that comparisons to \cite{berens2016znet}, \cite{liao2019evaluate}, \cite{method:deeplung}, and \cite{method:deepseed} are more fair and relevant given the proximity in the number of trainable parameters.

Moreover, to evaluate the impact of testing time augmentation (TTA), we compare the performance of our network with TTA against the network when no TTA is used. We find that TTA produces an enhanced 
% FROC score
CPM score
of 0.874, showing an increase of 0.3\% in contrast to the network when no TTA is used.
% being the best performing detector amongst all network approaches. 
This is in line with the finding in
% \cite{wang2021transbts,moshkov2020test,simonyan2014very,szegedy2016,shanmugam2020and}
\cite{wang2021transbts,moshkov2020test,simonyan2014very}
, in which TTA was demonstrated to be a simple and effective way to boost the performance of DL models.
% Accordingly, we continue using TTA as a part of our network. 
Fig. \ref{fig:nodule_detection_samples} compares pulmonary nodules as detected by our network against their correspondent ground-truth locations.

\begin{figure}[t]
\centering

\includegraphics[width=1.0\linewidth]{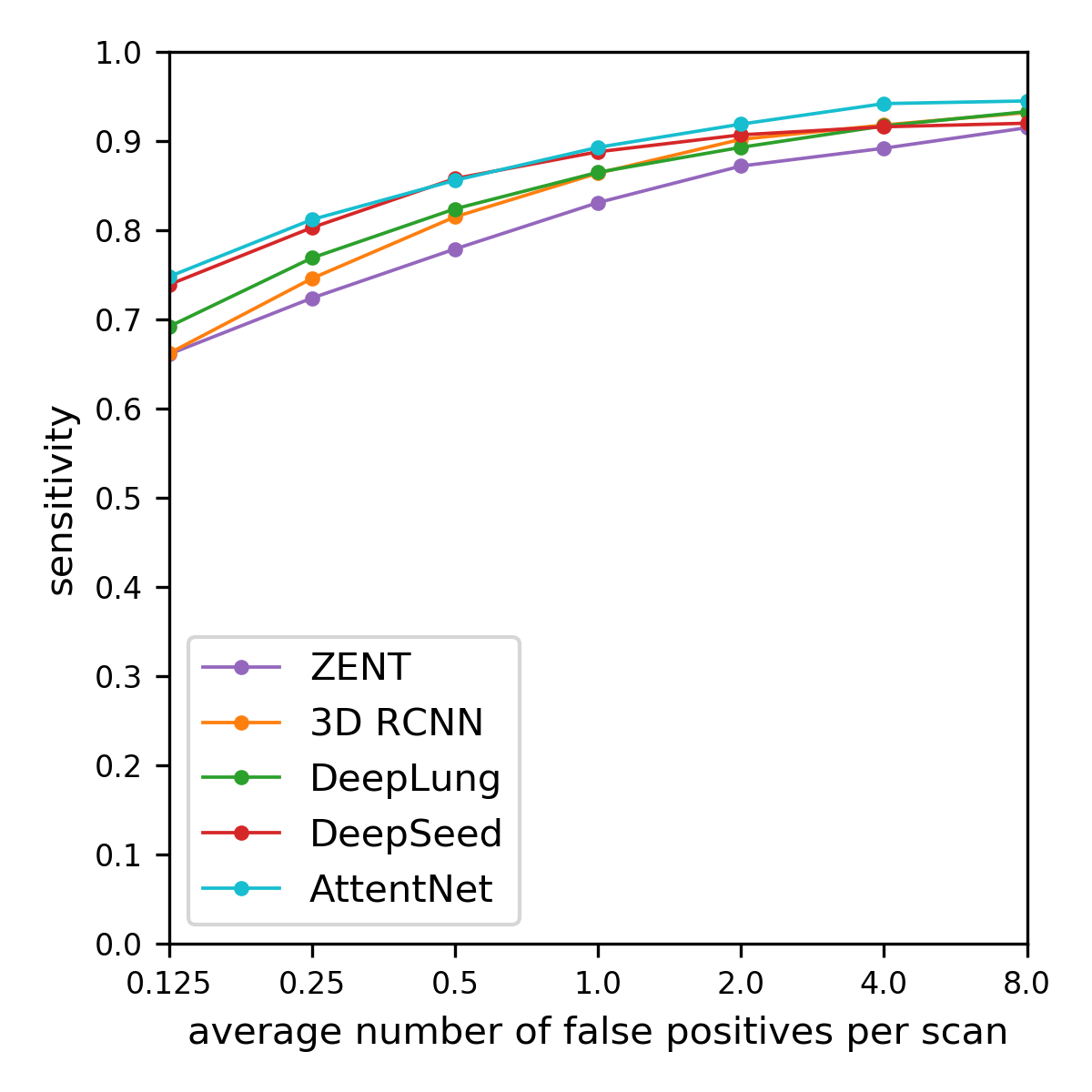}

\caption{
FROC of all systems under comparison using 10-folds cross-validation over LUNA16 dataset. Our proposed system produces the highest 
% FROC score
CPM score
(average sensitivity over all predefined false positive thresholds) of 0.874 with a total sensitivity of 95\%.
}

\label{fig:overall_froc_comparison}
\end{figure}

Overall, our proposed detector
and attention methods
demonstrate promising results within the pulmonary nodule detection task. AttentNet can efficiently achieve 
% state-of-the-art 
competitive
performance 
% with higher 
% FROC score
% CPM score
in contrast to 
% state-of-the-art
existing
detectors, 
% and
a total sensitivity of 95\%. See Fig. \ref{fig:overall_froc_comparison}.
AttentNet significantly outperforms \cite{berens2016znet}\cite{liao2019evaluate}, \cite{method:deeplung}\cite{method:deepseed} at 1.0, 2.0, and 4.0 false positives per scan, in which an effective clinical threshold is defined \cite{dataset:LUNA16}, and produces the highest sensitivity at the more challenging lower false positive per scan thresholds (i.e. 0.125, 0.25, and 0.5), while maintaining a compact network design with a relatively smaller number of learnable paramters.

\section{Conclusion} 
\label{section:conclusion}

% Pulmonary nodule detection is a challenging task due to the variable nodule morphology and sparsity of nodule locations within the lung region.
% 
In this work, 
% we present AttentNet, an ensemble 3D nodule detection framework of two stages (candidate proposal and false positive reduction), in which 
% we explore attention mechanisms
% 
% to assist CNNs in focusing on effective features and therby produce a more robust predictions.
we explore attention mechanisms to assist CNNs focus on essential features, thereby producing more robust predictions.
We propose a 3D cross-channel and a cross-sectional spatial attention unit
in which we demonstrate effectiveness within the 3D lung nodule detection task, using a two stage detector (candidate proposal and false positive reduction).
We show that by incorporating a fully convolutional networks, attention can be efficiently performed using richer descriptors of higher spatial dimensionality, improving the overall performance in contrast to popular multi-layer perceptron based attention networks while retaining model compactness.
Additionally, for the false positive reduction task,
we show that by jointly analysing inputs of different spatial
scales attention and using our zoom-in paths along with cross-channel, the network was able to aggregate information of different contextual levels and produce enhanced predictions. 
% 
% We also present convolutional zoom-in paths to assist the network in capturing spatial information from various semantic and spatial scales, we demonstrate the benefits of the proposed zoom-in path in the false positive reduction task.
% 
We carry out an extensive analysis on LUNA16 dataset and show the potential of incorporating fully convolutional attention within the 3D lung nodule detection context 
% AttentNet can outperform 
% state-of-the-art
% competing
comparing to baseline 
% lung nodule 
detectors where no attention is used
% a considerable margin, with 
% an FROC score
achieving a comparable CPM score of 0.874 and a total sensitivity of 0.95 while retaining a relatively small number of trainable parameters.

% \section{STATEMENT}
% \begin{itemize}
% \item Competing Interests (Not Applicable)
% \item Funding Information (Not Applicable)
% \item Author contribution
% \begin{itemize}
%     \item Majedaldein Almahasneh (lead): Conceptualization, Methodology, Data Collection, Software Development, Formal Analysis, Writing - Original Draft, Writing - Review and Editing.
    
%     \item Xianghua Xie (main supervisory and support): Conceptualization, Methodology, Supervision, Writing - Review and Editing.
    
%     \item Adeline Paiement (secondary supervisory and support): Conceptualization, Methodology, Supervision, Writing - Review and Editing.
% \end{itemize}
% \item Data Availability Statement (Not Applicable)
% \item Research Involving Human and/or Animals (Not Applicable)
% \item Informed Consent (Not Applicable)
% \end{itemize}

%%Harvard
% \bibliographystyle{model2-names.bst}\biboptions{authoryear}

% {\small \bibliography{refs}}

% References :
\bibliographystyle{apalike}
{\small
\bibliography{refs}}

\begin{thebibliography}{}

\bibitem[Al~Mohammad et~al., 2017]{al2017review}
Al~Mohammad, B., Brennan, P.~C., and Mello-Thoms, C. (2017).
\newblock A review of lung cancer screening and the role of computer-aided detection.
\newblock {\em Clinical radiology}, 72(6):433--442.

\bibitem[Armato~III et~al., 2011]{dataset:LIDC}
Armato~III, S.~G., McLennan, G., Bidaut, L., McNitt-Gray, M.~F., Meyer, C.~R., Reeves, A.~P., Zhao, B., Aberle, D.~R., Henschke, C.~I., Hoffman, E.~A., et~al. (2011).
\newblock The lung image database consortium (lidc) and image database resource initiative (idri): a completed reference database of lung nodules on ct scans.
\newblock {\em Medical physics}.

\bibitem[Berens et~al., 2016]{berens2016znet}
Berens, M., van~der Gugten, R., de~Kaste, M., Manders, J., and Zuidhof, G. (2016).
\newblock Znet-lung nodule detection.

\bibitem[Chen et~al., 2021]{chen2021transunet}
Chen, J., Lu, Y., Yu, Q., Luo, X., Adeli, E., Wang, Y., Lu, L., Yuille, A.~L., and Zhou, Y. (2021).
\newblock Transunet: Transformers make strong encoders for medical image segmentation.
\newblock {\em arXiv preprint arXiv:2102.04306}.

\bibitem[Chen et~al., 2020]{chen2020end}
Chen, Y., Cao, P., Dou, L., and Yang, J. (2020).
\newblock An end-to-end framework for pulmonary nodule detection and false positive reduction from ct images.
\newblock In {\em The Fourth International Symposium on Image Computing and Digital Medicine}, pages 156--162.

\bibitem[Chen et~al., 2017]{NIPS2017_dualpathnetworks}
Chen, Y., Li, J., Xiao, H., Jin, X., Yan, S., and Feng, J. (2017).
\newblock Dual path networks.
\newblock In Guyon, I., Luxburg, U.~V., Bengio, S., Wallach, H., Fergus, R., Vishwanathan, S., and Garnett, R., editors, {\em Advances in Neural Information Processing Systems}, volume~30. Curran Associates, Inc.

\bibitem[Clevert et~al., 2016]{clevert2015fast}
Clevert, D.-A., Unterthiner, T., and Hochreiter, S. (2016).
\newblock Fast and accurate deep network learning by exponential linear units ({ELUs}).

\bibitem[Ding et~al., 2017]{method:accurateluna}
Ding, J., Li, A., Hu, Z., and Wang, L. (2017).
\newblock Accurate pulmonary nodule detection in computed tomography images using deep convolutional neural networks.
\newblock In Descoteaux, M., Maier-Hein, L., Franz, A., Jannin, P., Collins, D.~L., and Duchesne, S., editors, {\em Medical Image Computing and Computer Assisted Intervention -- MICCAI 2017}. Springer International Publishing.

\bibitem[Dosovitskiy et~al., 2020]{dosovitskiy2020image}
Dosovitskiy, A., Beyer, L., Kolesnikov, A., Weissenborn, D., Zhai, X., Unterthiner, T., Dehghani, M., Minderer, M., Heigold, G., Gelly, S., et~al. (2020).
\newblock An image is worth 16x16 words: Transformers for image recognition at scale.
\newblock In {\em International Conference on Learning Representations}.

\bibitem[Dou et~al., 2016]{method:dou2016multilevel}
Dou, Q., Chen, H., Yu, L., Qin, J., and Heng, P.-A. (2016).
\newblock Multilevel contextual 3-d cnns for false positive reduction in pulmonary nodule detection.
\newblock {\em IEEE Transactions on Biomedical Engineering}, 64(7):1558--1567.

\bibitem[Friedman, 2001]{friedman2001greedyGBM}
Friedman, J.~H. (2001).
\newblock Greedy function approximation: a gradient boosting machine.
\newblock {\em Annals of statistics}, pages 1189--1232.

\bibitem[Glorot et~al., 2011]{glorot2011deep}
Glorot, X., Bordes, A., and Bengio, Y. (2011).
\newblock Deep sparse rectifier neural networks.
\newblock In {\em Proceedings of the fourteenth international conference on artificial intelligence and statistics}, pages 315--323. JMLR Workshop and Conference Proceedings.

\bibitem[GU et~al., ]{gu40retrieval}
GU, J., WANG, F., QI, Y., SUN, Z., TIAN, Z., and ZHANG, Y.
\newblock Retrieval method of pulmonary nodule images based on multi-scale convolution feature fusion.
\newblock {\em Journal of Computer Applications}, 40(2):561--565.

\bibitem[Haibo et~al., 2021]{haibo2021improved}
Haibo, L., Shanli, T., Shuang, S., and Haoran, L. (2021).
\newblock An improved yolov3 algorithm for pulmonary nodule detection.
\newblock In {\em 2021 IEEE 4th Advanced Information Management, Communicates, Electronic and Automation Control Conference (IMCEC)}, volume~4, pages 1068--1072. IEEE.

\bibitem[He et~al., 2015]{he2015delving}
He, K., Zhang, X., Ren, S., and Sun, J. (2015).
\newblock Delving deep into rectifiers: Surpassing human-level performance on imagenet classification.
\newblock In {\em Proceedings of the IEEE international conference on computer vision}, pages 1026--1034.

\bibitem[He et~al., 2016]{method:resnet}
He, K., Zhang, X., Ren, S., and Sun, J. (2016).
\newblock Deep residual learning for image recognition.
\newblock In {\em Proceedings of the IEEE conference on computer vision and pattern recognition}, pages 770--778.

\bibitem[Hu et~al., 2018]{hu2018squeeze}
Hu, J., Shen, L., and Sun, G. (2018).
\newblock Squeeze-and-excitation networks.
\newblock In {\em Proceedings of the IEEE conference on computer vision and pattern recognition}, pages 7132--7141.

\bibitem[Huang et~al., 2017a]{huang2017densely}
Huang, G., Liu, Z., Van Der~Maaten, L., and Weinberger, K.~Q. (2017a).
\newblock Densely connected convolutional networks.
\newblock In {\em Proceedings of the IEEE conference on computer vision and pattern recognition}, pages 4700--4708.

\bibitem[Huang et~al., 2017b]{Huang_2017}
Huang, J., Rathod, V., Sun, C., Zhu, M., Korattikara, A., Fathi, A., Fischer, I., Wojna, Z., Song, Y., Guadarrama, S., et~al. (2017b).
\newblock Speed/accuracy trade-offs for modern convolutional object detectors.
\newblock In {\em Proceedings of the IEEE conference on computer vision and pattern recognition}.

\bibitem[Huang and Hu, 2019]{huang2019using}
Huang, W. and Hu, L. (2019).
\newblock Using a noisy u-net for detecting lung nodule candidates.
\newblock {\em IEEE Access}, 7:67905--67915.

\bibitem[Jacobs et~al., 2016]{jacobs2016computer}
Jacobs, C., van Rikxoort, E.~M., Murphy, K., Prokop, M., Schaefer-Prokop, C.~M., and van Ginneken, B. (2016).
\newblock Computer-aided detection of pulmonary nodules: a comparative study using the public lidc/idri database.
\newblock {\em European radiology}, 26(7):2139--2147.

\bibitem[Kopuklu et~al., 2019]{kopuklu2019resource}
Kopuklu, O., Kose, N., Gunduz, A., and Rigoll, G. (2019).
\newblock Resource efficient 3d convolutional neural networks.
\newblock In {\em Proceedings of the IEEE/CVF International Conference on Computer Vision Workshops}, pages 0--0.

\bibitem[Larici et~al., 2017]{larici2017lung}
Larici, A.~R., Farchione, A., Franchi, P., Ciliberto, M., Cicchetti, G., Calandriello, L., Del~Ciello, A., and Bonomo, L. (2017).
\newblock Lung nodules: size still matters.
\newblock {\em European Respiratory Review}, 26(146).

\bibitem[Li and Fan, 2020]{method:deepseed}
Li, Y. and Fan, Y. (2020).
\newblock Deepseed: 3d squeeze-and-excitation encoder-decoder convolutional neural networks for pulmonary nodule detection.
\newblock In {\em IEEE 17th International Symposium on Biomedical Imaging (ISBI)}, pages 1866--1869. IEEE.

\bibitem[Liao et~al., 2019]{liao2019evaluate}
Liao, F., Liang, M., Li, Z., Hu, X., and Song, S. (2019).
\newblock Evaluate the malignancy of pulmonary nodules using the 3-d deep leaky noisy-or network.
\newblock {\em IEEE transactions on neural networks and learning systems}, 30(11):3484--3495.

\bibitem[Lin et~al., 2017]{lin2017focal}
Lin, T.-Y., Goyal, P., Girshick, R., He, K., and Doll{\'a}r, P. (2017).
\newblock Focal loss for dense object detection.
\newblock In {\em Proceedings of the IEEE international conference on computer vision}, pages 2980--2988.

\bibitem[Lu et~al., 2021]{lu2021multi}
Lu, X., Chang, E.~Y., Hsu, C.-n., Du, J., and Gentili, A. (2021).
\newblock Multi-classification study of the tuberculosis with 3d cbam-resnet and efficientnet.
\newblock In {\em CLEF2021 Working Notes. CEUR Workshop Proceedings, Bucharest, Romania, CEUR-WS. org< http://ceur-ws. org>(September 21-24 2021)}.

\bibitem[Maas et~al., 2013]{maas2013rectifier}
Maas, A.~L., Hannun, A.~Y., Ng, A.~Y., et~al. (2013).
\newblock Rectifier nonlinearities improve neural network acoustic models.
\newblock In {\em Proc. icml}, volume~30, page~3. Citeseer.

\bibitem[Matsumoto et~al., 2013]{matsumoto2013computer}
Matsumoto, S., Ohno, Y., Aoki, T., Yamagata, H., Nogami, M., Matsumoto, K., Yamashita, Y., and Sugimura, K. (2013).
\newblock Computer-aided detection of lung nodules on multidetector ct in concurrent-reader and second-reader modes: a comparative study.
\newblock {\em European journal of radiology}, 82(8):1332--1337.

\bibitem[Moshkov et~al., 2020]{moshkov2020test}
Moshkov, N., Mathe, B., Kertesz-Farkas, A., Hollandi, R., and Horvath, P. (2020).
\newblock Test-time augmentation for deep learning-based cell segmentation on microscopy images.
\newblock {\em Scientific reports}, 10(1):1--7.

\bibitem[Nawshad et~al., 2021]{nawshad2021attention}
Nawshad, M.~A., Shami, U.~A., Sajid, S., and Fraz, M.~M. (2021).
\newblock Attention based residual network for effective detection of covid-19 and viral pneumonia.
\newblock In {\em 2021 International Conference on Digital Futures and Transformative Technologies (ICoDT2)}, pages 1--7. IEEE.

\bibitem[Park et~al., 2018]{park2018bam}
Park, J., Woo, S., Lee, J.-Y., and Kweon, I.-S. (2018).
\newblock Bam: Bottleneck attention module.
\newblock In {\em British Machine Vision Conference (BMVC)}. British Machine Vision Association (BMVA).

\bibitem[Pearl, 2014]{pearl2014probabilistic}
Pearl, J. (2014).
\newblock {\em Probabilistic reasoning in intelligent systems: networks of plausible inference}.
\newblock Elsevier.

\bibitem[Polat et~al., 2018]{polat2018false}
Polat, G., Halici, U., and Dogrusoz, Y.~S. (2018).
\newblock False positive reduction in lung computed tomography images using convolutional neural networks.
\newblock {\em arXiv preprint arXiv:1811.01424}.

\bibitem[Ren et~al., 2015]{ren2015faster}
Ren, S., He, K., Girshick, R., and Sun, J. (2015).
\newblock Faster r-cnn: Towards real-time object detection with region proposal networks.
\newblock {\em Advances in neural information processing systems}, 28:91--99.

\bibitem[Riquelme and Akhloufi, 2020]{riquelme2020deep}
Riquelme, D. and Akhloufi, M.~A. (2020).
\newblock Deep learning for lung cancer nodules detection and classification in ct scans.
\newblock {\em AI}, 1(1):28--67.

\bibitem[Ronneberger et~al., 2015]{ronneberger2015u}
Ronneberger, O., Fischer, P., and Brox, T. (2015).
\newblock U-net: Convolutional networks for biomedical image segmentation.
\newblock In {\em International Conference on Medical image computing and computer-assisted intervention}, pages 234--241. Springer.

\bibitem[Sangeroki and Cenggoro, 2021]{sangeroki2021fast}
Sangeroki, B.~A. and Cenggoro, T.~W. (2021).
\newblock A fast and accurate model of thoracic disease detection by integrating attention mechanism to a lightweight convolutional neural network.
\newblock {\em Procedia Computer Science}, 179:112--118.

\bibitem[Setio et~al., 2017]{dataset:LUNA16}
Setio, A. A.~A., Traverso, A., De~Bel, T., Berens, M.~S., Van Den~Bogaard, C., Cerello, P., Chen, H., Dou, Q., Fantacci, M.~E., Geurts, B., et~al. (2017).
\newblock Validation, comparison, and combination of algorithms for automatic detection of pulmonary nodules in computed tomography images: the luna16 challenge.
\newblock {\em Medical image analysis}.

\bibitem[Shen et~al., 2021]{shen2021classification}
Shen, L.-H., Wang, X.-H., Gao, M.-X., and Li, B. (2021).
\newblock Classification of benign-malignant pulmonary nodules based on multi-view improved dense network.
\newblock In {\em International Conference on Intelligent Computing}, pages 582--593. Springer.

\bibitem[Shi, 2018]{shi2018lung}
Shi, J. (2018).
\newblock Lung nodule detection using convolutional neural networks.
\newblock {\em Electrical Engineering and Computer Sciences. Berkeley, California: University of California at Berkeley}.

\bibitem[Shrivastava et~al., 2016]{method:OHEM}
Shrivastava, A., Gupta, A., and Girshick, R. (2016).
\newblock Training region-based object detectors with online hard example mining.
\newblock In {\em Proceedings of the IEEE conference on computer vision and pattern recognition}, pages 761--769.

\bibitem[Simonyan and Zisserman, 2014]{simonyan2014very}
Simonyan, K. and Zisserman, A. (2014).
\newblock Very deep convolutional networks for large-scale image recognition.
\newblock {\em arXiv preprint arXiv:1409.1556}.

\bibitem[Soviany and Ionescu, 2018]{paper:opttradeoff}
Soviany, P. and Ionescu, R. (2018).
\newblock Optimizing the trade-off between single-stage and two-stage deep object detectors using image difficulty prediction.
\newblock In {\em SYNASC}. IEEE.

\bibitem[Sun et~al., 2021]{sun2021attention}
Sun, L., Wang, Z., Pu, H., Yuan, G., Guo, L., Pu, T., and Peng, Z. (2021).
\newblock Attention-embedded complementary-stream cnn for false positive reduction in pulmonary nodule detection.
\newblock {\em Computers in Biology and Medicine}, 133:104357.

\bibitem[Tan et~al., 2018]{tan2018fast}
Tan, J., Huo, Y., Liang, Z., and Li, L. (2018).
\newblock A fast automatic juxta-pleural lung nodule detection framework using convolutional neural networks and vote algorithm.
\newblock In {\em International Workshop on Patch-based Techniques in Medical Imaging}, pages 85--92. Springer.

\bibitem[Tang et~al., 2018]{tang2018automated}
Tang, H., Kim, D.~R., and Xie, X. (2018).
\newblock Automated pulmonary nodule detection using 3d deep convolutional neural networks.
\newblock In {\em 2018 IEEE 15th International Symposium on Biomedical Imaging (ISBI 2018)}, pages 523--526. IEEE.

\bibitem[Vaswani et~al., 2017]{vaswani2017attention}
Vaswani, A., Shazeer, N., Parmar, N., Uszkoreit, J., Jones, L., Gomez, A.~N., Kaiser, {\L}., and Polosukhin, I. (2017).
\newblock Attention is all you need.
\newblock In {\em Advances in neural information processing systems}, pages 5998--6008.

\bibitem[Wang et~al., 2019]{wang2019lung}
Wang, Q., Shen, F., Shen, L., Huang, J., and Sheng, W. (2019).
\newblock Lung nodule detection in ct images using a raw patch-based convolutional neural network.
\newblock {\em Journal of digital imaging}, 32(6):971--979.

\bibitem[Wang et~al., 2021]{wang2021transbts}
Wang, W., Chen, C., Ding, M., Yu, H., Zha, S., and Li, J. (2021).
\newblock Transbts: Multimodal brain tumor segmentation using transformer.
\newblock In {\em International Conference on Medical Image Computing and Computer-Assisted Intervention}, pages 109--119. Springer.

\bibitem[Woo et~al., 2018]{woo2018cbam}
Woo, S., Park, J., Lee, J.-Y., and Kweon, I.~S. (2018).
\newblock Cbam: Convolutional block attention module.
\newblock In {\em Proceedings of the European conference on computer vision (ECCV)}, pages 3--19.

\bibitem[Wo{\'z}niak et~al., 2018]{wozniak2018small}
Wo{\'z}niak, M., Po{\l}ap, D., Capizzi, G., Sciuto, G.~L., Ko{\'s}mider, L., and Frankiewicz, K. (2018).
\newblock Small lung nodules detection based on local variance analysis and probabilistic neural network.
\newblock {\em Computer methods and programs in biomedicine}, 161:173--180.

\bibitem[Xie et~al., 2017]{method:resnext}
Xie, S., Girshick, R., Doll{\'a}r, P., Tu, Z., and He, K. (2017).
\newblock Aggregated residual transformations for deep neural networks.
\newblock In {\em Proceedings of the IEEE conference on computer vision and pattern recognition}, pages 1492--1500.

\bibitem[Xie, 2017]{Xie2017}
Xie, Z. (2017).
\newblock 3d region proposal u-net with dense and residual learning for lung nodule detection.
\newblock Accessed: 2024-07-11.

\bibitem[Yu et~al., 2020]{yu2020amplification}
Yu, M., Cai, L., Gao, L., and Gao, J. (2020).
\newblock Amplification method of lung nodule data based on dcgan generation algorithm.
\newblock In {\em International Conference of Pioneering Computer Scientists, Engineers and Educators}, pages 563--576. Springer.

\bibitem[Yuan et~al., 2021]{yuan2021tokens}
Yuan, L., Chen, Y., Wang, T., Yu, W., Shi, Y., Jiang, Z., Tay, F.~E., Feng, J., and Yan, S. (2021).
\newblock Tokens-to-token vit: Training vision transformers from scratch on imagenet.
\newblock {\em arXiv preprint arXiv:2101.11986}.

\bibitem[Zhang et~al., 2019]{zhang2019classification}
Zhang, G., Yang, Z., Gong, L., Jiang, S., and Wang, L. (2019).
\newblock Classification of benign and malignant lung nodules from ct images based on hybrid features.
\newblock {\em Physics in Medicine \& Biology}, 64(12):125011.

\bibitem[Zhao et~al., 2003]{zhao2003automatic}
Zhao, B., Gamsu, G., Ginsberg, M.~S., Jiang, L., and Schwartz, L.~H. (2003).
\newblock Automatic detection of small lung nodules on ct utilizing a local density maximum algorithm.
\newblock {\em journal of applied clinical medical physics}, 4(3):248--260.

\bibitem[Zheng et~al., 2021]{zheng2021rethinking}
Zheng, S., Lu, J., Zhao, H., Zhu, X., Luo, Z., Wang, Y., Fu, Y., Feng, J., Xiang, T., Torr, P.~H., et~al. (2021).
\newblock Rethinking semantic segmentation from a sequence-to-sequence perspective with transformers.
\newblock In {\em Proceedings of the IEEE/CVF Conference on Computer Vision and Pattern Recognition}, pages 6881--6890.

\bibitem[Zhu et~al., 2018]{method:deeplung}
Zhu, W., Liu, C., Fan, W., and Xie, X. (2018).
\newblock Deeplung: Deep 3d dual path nets for automated pulmonary nodule detection and classification.
\newblock In {\em IEEE Winter Conference on Applications of Computer Vision (WACV)}, pages 673--681.

\end{thebibliography}

\end{document}